\documentclass{article}

\usepackage[preprint]{neurips_2025}
\usepackage{graphicx}
\usepackage{multirow} 
\usepackage{wrapfig}  % 用于浮动图片
\usepackage{enumitem}
\usepackage{amsmath}    % 最常用的数学宏包，提供多种数学环境
\usepackage{amssymb}    % 提供更多数学符号
\usepackage{mathtools}  % amsmath 的扩展，提供更多功能
\usepackage{rotating} % For rotating text
\usepackage{booktabs} % For better-looking horizontal lines
\usepackage{tabularx} % For flexible table widths

\usepackage[utf8]{inputenc} % allow utf-8 input
\usepackage[T1]{fontenc}    % use 8-bit T1 fonts
\usepackage{hyperref}       % hyperlinks
\usepackage{url}            % simple URL typesetting
\usepackage{booktabs}       % professional-quality tables
\usepackage{amsfonts}       % blackboard math symbols
\usepackage{nicefrac}       % compact symbols for 1/2, etc.
\usepackage{microtype}      % microtypography
\usepackage{xcolor}         % colors

\title{Teach2Eval: An Indirect Evaluation Method for LLM by Judging How It Teaches}

\author{
\textbf{Yuhang Zhou}\(^{1,2*}\) \,
\textbf{Xutian Chen}\(^{1*}\) \,
\textbf{Yixin Cao}\(^{1~\dagger}\) \,
\textbf{Yuchen Ni}\(^{1,2}\) \,
\textbf{Yu He}\(^{1,2}\) \\
\textbf{Siyu Tian}\(^{1}\) \,
\textbf{Xiang Liu}\(^{3}\) \,
\textbf{Jian Zhang}\(^{4}\) \,
\textbf{Chuanjun Ji}\(^{4}\) \,
\textbf{Guangnan Ye}\(^{1,2~\dagger}\) \,
\textbf{Xipeng Qiu}\(^{1,2}\)
\\
\(^{1}\)School of Computer Science, Fudan University \\
\(^{2}\)Shanghai Innovation Institute \\
\(^{3}\)Computer Science Department, NYU Shanghai \\
\(^{4}\)DataGrand Inc. \\
\\
}

\begin{document}

\maketitle

\renewcommand{\thefootnote}{\fnsymbol{footnote}} % 使用符号作为脚注标记
\footnotetext[1]{Contribute equally to this work.}
\footnotetext[2]{The corresponding authors:Yixin Cao(yxcao@fudan.edu.cn), Guangnan Ye(yegn@fudan.edu.cn) }

\renewcommand{\thefootnote}{\arabic{footnote}} % 恢复默认的脚注编号格式

\begin{abstract}
  % The abstract paragraph should be indented \nicefrac{1}{2}~inch (3~picas) on
  % both the left- and right-hand margins. Use 10~point type, with a vertical
  % spacing (leading) of 11~points.  The word \textbf{Abstract} must be centered,
  % bold, and in point size 12. Two line spaces precede the abstract. The abstract
  % must be limited to one paragraph.
  % Recent advances in large language models (LLMs) have expanded their capabilities, but evaluation methods have not kept pace. Traditional direct evaluation often relies on task-specific metrics that suffer from data contamination and reliability issues. To address these, we propose Teach2Eval, a scalable, multi-dimensional evaluation inspired by the Feynman Technique. Unlike traditional methods that directly assess LLMs on datasets, Teach2Eval uses LLMs to guide several weaker "student" models, and the overall performance improvement of weak models serves as an indirect metric of the LLM's capabilities. We apply this method across 60 datasets, covering Knowledge, Reasoning, Understanding, and Multilingual tasks, with 26 state-of-the-art models. Our results show that Teach2Eval aligns with existing evaluation platforms while offering a more cost-effective, accurate measure of open-ended performance. Additionally, the method is robust under scaling, confirming the validity of the Scaling Law in our framework.
  Recent progress in large language models (LLMs) has outpaced the development of effective evaluation methods. Traditional benchmarks rely on task-specific metrics and static datasets, which often suffer from fairness issues, limited scalability, and contamination risks. In this paper, we introduce Teach2Eval , an indirect evaluation framework inspired by the Feynman Technique. Instead of directly testing LLMs on predefined tasks, our method evaluates a model’s multiple abilities to teach weaker "student" models to perform tasks effectively. By converting open-ended tasks into standardized multiple-choice questions (MCQs) through teacher-generated feedback, Teach2Eval enables scalable, automated, and multi-dimensional assessment. Our approach not only avoids data leakage and memorization but also captures a broad range of cognitive abilities that are orthogonal to current benchmarks. Experimental results across 26 leading LLMs show strong alignment with existing human and model-based dynamic rankings, while offering additional interpretability for training guidance.

% \textit{If you can't explain it simply, you don't understand it well enough. ---- Albert Einstein}
\end{abstract}

\section{Introduction}
% Progress in large language models (LLMs) has considerably broadened their capabilities, moving past conventional NLP approaches to address diverse general-purpose tasks~\cite{openai2023gpt4,touvron2023llama}. However, progress in model evaluation has been relatively slow. Model performance is mainly assessed through performance metrics on specific task datasets, such as GSM8K~\cite{cobbe2021training}, MATH~\cite{hendrycks2021measuring}, MMLU~\cite{hendrycks2020measuring}, and BigBench~\cite{srivastava2022beyond}. These datasets are prone to issues like data contamination and the challenges of constructing high-quality data, raising concerns about the validity of the evaluations~\cite{wei2023skywork,zhou2023don}. Additionally, direct evaluations fail to capture subtle differences and the diversity of capabilities.

Recent advances in large language models (LLMs) have significantly expanded their capabilities, enabling them to tackle a diverse range of general-purpose tasks beyond traditional natural language processing (NLP) applications~\cite{openai2023gpt4,touvron2023llama}. However, the progress in model evaluation has not kept pace. Evaluation has primarily relied on task-specific performance metrics, such as those from GSM8K~\cite{cobbe2021training}, MATH~\cite{hendrycks2021measuring}, MMLU~\cite{hendrycks2020measuring}, and BigBench~\cite{srivastava2022beyond}. 
These datasets are often prone to various fairness issues~\cite{wei2023skywork,zhou2023don}, e.g., \textit{the testing data covers sufficient capabilities of interest in practice?} \textit{how to fairly judge the open-ended responses?}
%These datasets are often prone to data contamination, such as issues with overly tailored prompts, leaks in the training data, and the challenges of constructing high-quality benchmarks. This raises concerns about the validity and reliability of current evaluation approaches~\cite{wei2023skywork,zhou2023don}. 

% Moreover, traditional evaluations often fail to capture subtle differences and the full spectrum of model capabilities.

\begin{figure}[]
\centerline{\includegraphics[width=0.85\linewidth]{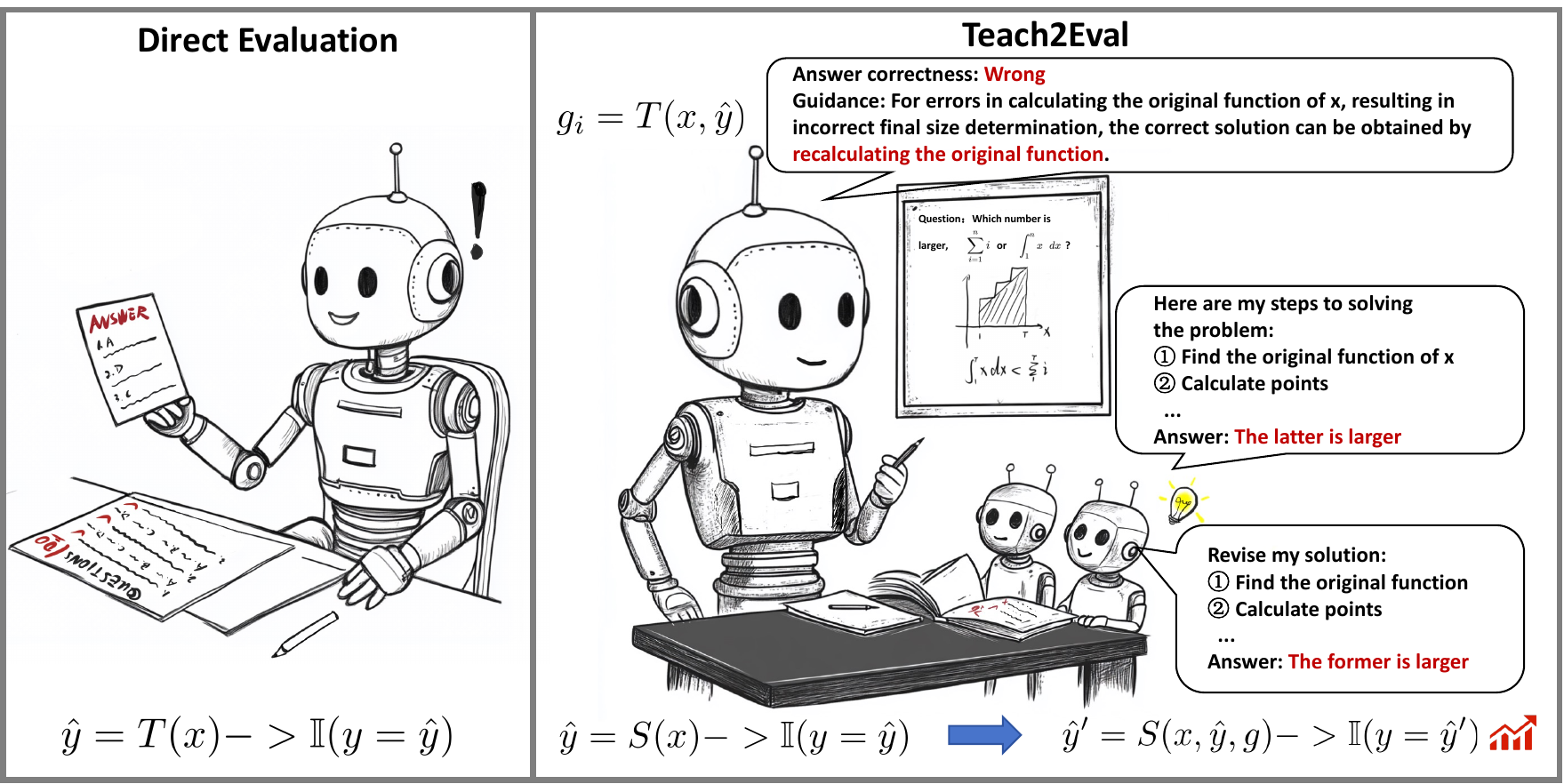}}
\caption{Comparing two evaluation methods. The left shows the use of static benchmarks to directly evaluate LLM, which may lead to data contamination and result in artificially high scores; On the right is Teach2Eval, which generates guidance to enhance the ability of the weak model as an indirect metric for LLM evaluation.}
\label{img:irrationality_example}
\end{figure}

In this paper, we propose Teach2Eval, an indirect evaluation method that targets scalable, quantifiable, and multi-dimensional evaluation of LLMs. Inspired by the Feynman Technique~\cite{feynman2018feynman}, which states "explaining or teaching concepts in a simple, concise manner deepens one's understanding of the material", we introduce this principle as a tool for evaluating abilities beyond learning strategy. Unlike traditional evaluation methods, Teach2Eval does not directly assess LLMs through performance metrics on given datasets. Instead, we leverages the target to teach weak "student" models to complete tasks, as shown in Figure~\ref{img:irrationality_example}, and the overall performance gains of students shall reflect multi-dimensional capabilities of the teacher. That is, better students, better teacher LLMs.

There are several benefits of the above idea, which guides the design of such an indirect method.
First, it transforms open-ended questions into multi-choice questions (MCQs) while preserving task difficulty, enabling more standardized and scalable assessment. As LLMs encounter increasing open-ended queries, traditional evaluation methods face challenges due to reliance on costly human judgment or inconsistent automated metrics~\cite{gu2024survey, li2024generation}. While direct conversion to MCQs simplifies evaluation, it often fails to capture the generative capabilities of LLMs.
In contrast, Teach2Eval leverages LLMs to generate feedback that guides weaker models to answer in MCQ format. We construct a benchmark of 60 datasets across four categories—Knowledge, Reasoning, Understanding, and Multilingual—and standardize them as MCQs with added misleading options derived from original answers. Evaluation is based on student model performance gains after teacher-guided learning.
Importantly, the teacher model receives no access to MCQ choices, ensuring that evaluation reflects student accuracy rather than exposure to answer options. This approach enables cost-effective, fully automated evaluation for open-ended tasks.
% First, we can convert open-ended questions into Multi-Choice questions (MCQs) while maintaining the difficulty, where the latter is clearly easy to measure.
% Along with an increasing number of users, LLMs are tasked with more and more open-ended queries. However, previous evaluation methods have struggled due to the lack of standardized measurements, often relying on costly human evaluation or LLM-as-a-Judge~\cite{gu2024survey,li2024generation}. 
% A straightforward solution is to convert open tasks to MCQs. While, this not only simplifies the problem, but also hardly evaluate the generation ability (more like verification), leading to a failure in reflecting the true capabilities of LLMs.
% While, for Teach2Eval, the target LLMs are maintained to generate feedbacks, which are used to teach and guide weak models to answer in the format of MCQs.
% Therefore, we compile a collection of 60 datasets, divide into four key tasks: Knowledge, Reasoning, Understanding, and Multilingual. We then unify their format as MCQs and complement with misleading choices (correct choices are given in the original datasets).
% The metric is then defined based on performance gains of students after the teacher's guidance.
% Note that the teacher LLM is not given the MCQ options, no matter correct or misleading ones, ensuring evaluation is based on the student models' response accuracy. This provides a more cost-effective and automated measurement for open scenarios.

The second benefit of indirect evaluation is the involvement of comprehensive abilities.
%As AI advances toward superhuman intelligence, evaluating such models will become more complex. 
Traditional benchmarks usually focus on one or several targeted task-specific capabilities through the design of questions~\cite{cao2025toward}. While, Teach2Eval does not rely on the questions themselves (orthogonal to existing benchmarks). It grounds the evaluation in practice, and the teaching process naturally involves several practical abilities.
%will become too simple to differentiate true capabilities. Some studies also suggest weak models struggle to judge strong ones directly~\cite{khan2024debating}. Our approach avoids this by focusing on the improvement of weak models through LLM guidance.
We categorize them into four levels based on Bloom’s taxonomy~\cite{krathwohl2002revision}: Application, Judgment, Guidance, and Reflection.
Based on this, we design a multi-dimensional metric, and also prove that such a metric well supports the claims above.

% 上一句可能也要改

The third benefit is addressing the problem of data contamination. This interactive evaluation mechanism introduces dynamic randomness, and requires models to possess a deeper understanding of the knowledge in order to effectively teach the student model, memorizing answers becomes ineffective under this setup. To validate this, we designed a series of experiments based on the multi-dimensional cognitive abilities, demonstrating that Teach2Eval not only mitigates data contamination issue, but also provides fine-grained guidance for model training and development.

% 第三个好处是Teach2Eval能够解决传统评测存在的数据泄露问题，这种交互的评测机制一方面引入了动态随机性，另一方面模型需要对知识有更深的理解才能够教会学生模型，记住答案的方式在此评测机制上失效。对此，我们设计了一系列的实验，基于上述所提及的多维认知能力，验证了Teach2Eval在解决数据泄露问题的基础上，能够为模型训练与进化提供一系列细粒度的指导。

To evaluate this approach, we test 26 latest LLMs and the results show our method’s results align closely with platforms like Chatbot Arena~\cite{chiang2024chatbot}, LiveBench~\cite{white2024livebench}, and GPT-as-a-Judge, with similarities of 0.91, 0.92, and 0.88, respectively. Note that our proposed evaluation is automated and cost-effective.Also, we evaluate the convergence and robustness of this method based on a random combination of student models and longer interaction turns. We present evidence and advantages of using this evaluation method to guide model training in the final, for instance, by analyzing the evolution of the two capabilities during model training, our method provides clear directions for preventing model overfitting.

% Also, student models show stable improvement after a few iterations under LLM guidance, demonstrating the stability of Teach2Eval. We also evaluate reasoning models, which showes significant improvements across all tasks. Additionally, our results confirm that the Scaling Law~\cite{kaplan2020scaling} holds under our evaluation method, and the method is not sensitive to the choice of weak models.

We make the following contributions:

\begin{itemize}[itemsep=0pt, leftmargin=*]
\item We propose a novel, general indirect evaluation method Teach2Eval, which effectively addresses multiple issues of current evaluation approaches. We have open-sourced our code and data at https://github.com/zhiqix/Teach2Eval.

\item We define the metric based on students' improvements, and demonstrates its effectiveness in reflecting several cognitive abilities of the teacher LLMs, these abilities are orthogonal to the current benchmarks.

\item Our approach not only provides an effective overall ranking of models at low cost, but also enables fine-grained capability analysis to guide model training and refinement.
\end{itemize}

% 我们的方法不仅能够在低成本的情况下提供有效的模型群体综合排名，同时细粒度的能力分析能够进一步为模型的训练与进化指导方向。

\section{Related Work}

\subsection{Benchmarks}
% The most prevalent benchmarks are static, groundtruth-based ones, typically in the form of MCQs or question-answering (QA) tasks with predefined answers and test cases. These benchmarks can be categorized according to the 3H criteria~\cite{bai2022training}, and in our study, we focus primarily on helpful benchmarks. These benchmarks cover a range of topics and can be divided into four categories: Understanding, Knowledge, Reasoning, and Multilingual, as shown in Figure~\ref{img:dataset} in the Appendix~\ref{app:dataset}. Notable examples in this category include MMLU~\cite{hendrycks2020measuring}, GSM-8K~\cite{hendrycks2021measuring}, BigBench~\cite{srivastava2022beyond}, AGIEval~\cite{zhong2023agieval}, and HellaSwag~\cite{zellers2019hellaswag}. 

% However, static benchmarks come with certain limitations, including contamination, saturation, overfitting, and a lack of human alignment~\cite{yang2023rethinking,oren2023proving}. DynaBench~\cite{kiela2021dynabench} identified these challenges and proposed the use of real-time benchmarks that integrate human input into the cycle, building on classical NLP benchmark methods. Our system adopts a similar spirit, introducing perturbations from multiple weak models to increase the difficulty of answering, which can be generalized to a wide range of benchmarks.

The most prevalent benchmarks for evaluating models are static, groundtruth-based ones, such as MCQs or QA tasks with predefined answers. These benchmarks can be categorized according to the 3H criteria~\cite{bai2022training}, with a focus on helpful benchmarks. They cover a variety of topics, including Understanding, Knowledge, Reasoning, and Multilingual tasks, as illustrated in Figure~\ref{img:dataset} of Appendix~\ref{app:dataset}. Examples include MMLU~\cite{hendrycks2020measuring}, GSM-8K~\cite{hendrycks2021measuring}, BigBench~\cite{srivastava2022beyond}, AGIEval~\cite{zhong2023agieval}, and HellaSwag~\cite{zellers2019hellaswag}.

However, static benchmarks have limitations like contamination, saturation, overfitting, and a lack of human alignment~\cite{yang2023rethinking, oren2023proving}. DynaBench~\cite{kiela2021dynabench} addressed these issues by introducing real-time benchmarks that incorporate human input. Our system adopts a similar approach, using perturbations from weak models to increase task difficulty, which can be extended to a variety of benchmarks.

\begin{wrapfigure}{r}{0.5\textwidth} % 右对齐，宽度为页面宽度的一半
    \centering
    \includegraphics[width=0.48\textwidth]{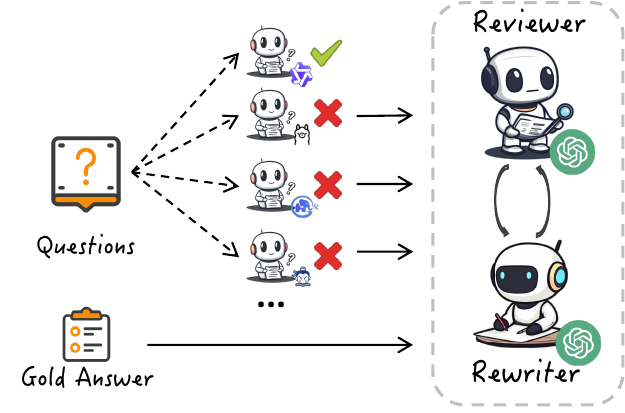}
    \caption{Use multiple weak models to answer questions and collect incorrect options with gold answers. Use GPT-4o as a rewriter and reviewer to convert the original question into MCQ format.}
    \label{fig:data_construct}
\end{wrapfigure}

\subsection{LLM Evaluation}

Current LLM evaluation methods can be classified into three categories. First, automatic evaluation using metrics is common, with benchmarks like MCQ-based datasets using accuracy, sentiment analysis datasets using F1 score, and text generation datasets using BertScore~\cite{zhang2019bertscore}. These metrics provide quick performance assessments but are limited to simple measures. Second, manual evaluation through human judgment involves methods like AdaTest ~\cite{chen2023adatest}, which gather human feedback, and approaches like DynaBench~\cite{kiela2021dynabench} and Chatbot Arena~\cite{chiang2024chatbot} that use crowdsourcing for diverse evaluations. Third, LLM-as-a-Judge uses LLMs for evaluation, such as MT-Bench~\cite{zheng2023judging}, which compares model labels to human labels, and other approaches that explore whether weak models can assess stronger ones~\cite{khan2024debating}. Currently, it is crucial to address the issue of generalization in evaluation~\cite{}, our method focuses on weak models’ improvement to indirect measure the multi-dimensional abilities of LLMs, ensuring accuracy, economy, and scalability.

\begin{figure*}[ht]
\centerline{\includegraphics[width=\textwidth]{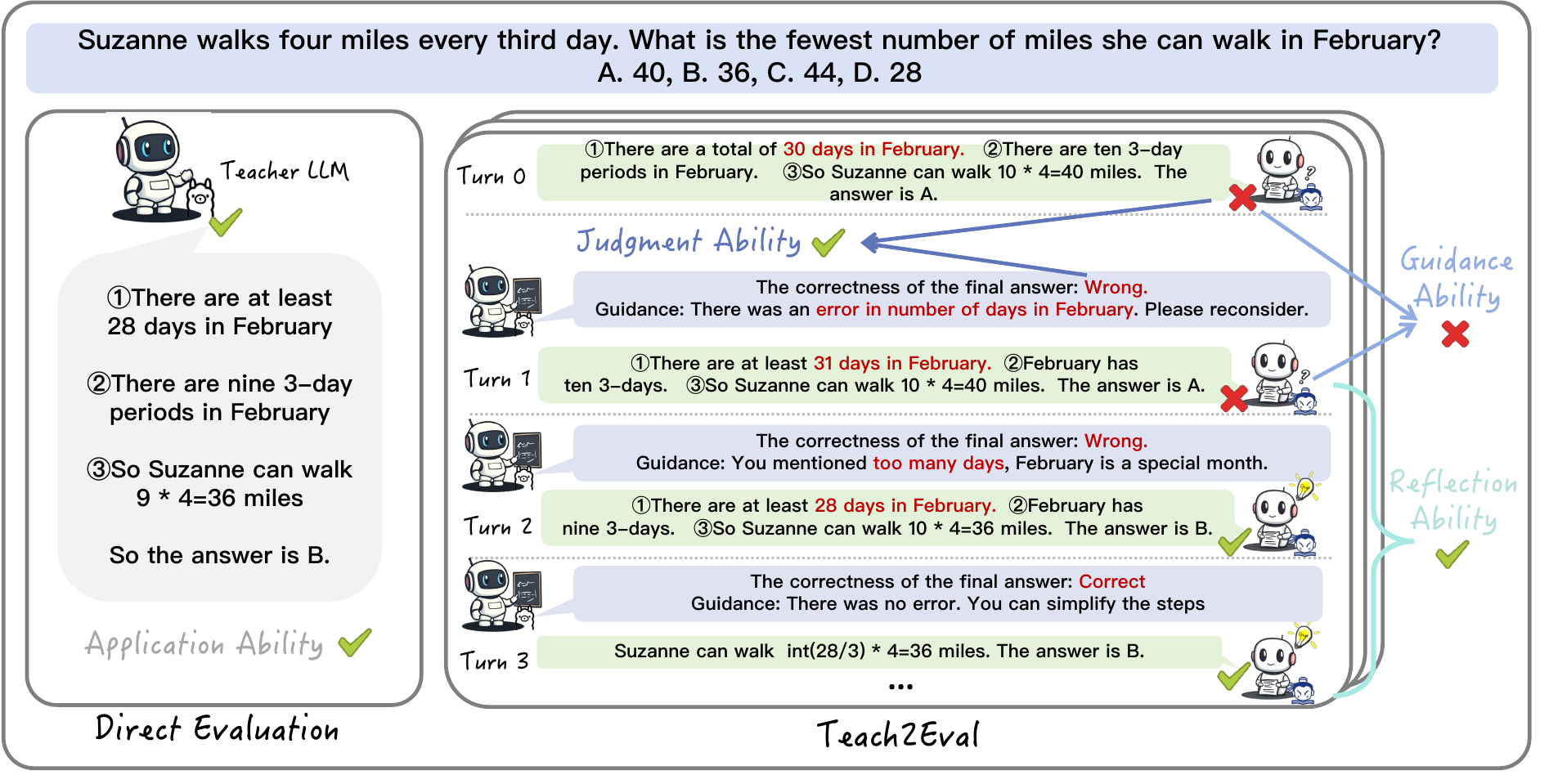}}
\caption{The overall design of Teach2Eval. On the left, Application Ability is obtained through direct evaluation, while on the right, multiple turns of guidance from the teacher LLM are used to guide the weak student model. The final performance gains of the student models is used as an indirect metirc of the LLM's ability. At the same time, by measuring the changes in student model’s ability between dialogues and the accuracy of teacher model judgments, various cognitive abilities of teacher models can be measured.}
\label{img:main}
\end{figure*}

\section{Method}

\subsection{Data Construction}

% \begin{figure}[]
% \centerline{\includegraphics[width=0.7\linewidth]{fig/data_construct.pdf}}
% \caption{Use multiple weak models to answer questions and collect incorrect options with gold answers. Use GPT-4o as a rewriter and reviewer to convert the original question into MCQ format.}
% \label{img:data_construct}
% \end{figure}

To ensure the effectiveness and diversity of the evaluation, we collect 60 datasets, and categorize them into Knowledge, Reasoning, Understanding, and Multilingual domains, with detailed information provided in the Appendix~\ref{app:dataset}. Specifically, due to the limited abilities of weak models, it is challenging to extract information from their open-ended responses, the MCQ format is better suited for accurately evaluating the capabilities of weak models. Therefore, we convert all the data into MCQ format. Specifically, to construct the set of options, we utilize a group of models to answer the questions and collect incorrect answers, as shown in Figure~\ref{fig:data_construct}. With these answers and the gold answer from datasets, we use the GPT-4o as the rewriter and reviewer to ensure proper formatting across the dataset. Besides, to examine LLM's alibities across various difficulty levels, we leverage the Qwen-family~\cite{yang2024qwen2} models of different parameter sizes to sequentially answer all questions. Based on the accuracy of the answers provided by different model sizes, we categorize the questions by five difficulty levels, detailed information can be found in Appendix~\ref{app:dataset}.

\subsection{Teach2Eval}

We propose \textit{Teach2Eval}, an indirect evaluation framework that assesses multi-dimensional capabilities of LLMs by measuring their ability to guide weak student models toward improved performance, as shown in Figure~\ref{img:main}.

In this setup, each student model $ S_j $ first answers a question $ d_i $ from dataset $ D $, using only the question and options. The teacher LLM $ T $ then evaluates and provides guidance without seeing the answer choices, ensuring open-ended problem-solving. In subsequent rounds, the student model refines its answer based on the latest guidance $ g_{j,i,t} $ and its prior response $ a_{j,i,t-1} $. The teacher, however, receives the full interaction history to support reflective reasoning.

The student model’s response at iteration $ t $ is defined as:

\begin{equation}
S_{j,i,t} =
\begin{cases}
S_{j,i,0}(d_i), & \text{if } t = 0, \\
S_{j,i,t}(d_i, a_{j,i,t-1}, g_{j,i,t}), & \text{if } t > 0.
\end{cases}
\end{equation}

The teacher LLM's output includes both judgment $ j_{j,i,t} $ and guidance $ g_{j,i,t} $, formulated as:

\begin{equation}
j_{j,i,t}, g_{j,i,t} = T \left( d_i, \left\{ a_{j,i,0}, \dots, g_{j,i,t-2},a_{j,i,t-1} \right\} \right)
\end{equation}

Let $ P(S) $ denote the accuracy of the student model. We define $ \Delta P_t(S) $ as the change in accuracy between iterations:

\begin{equation}
\Delta P_t(S) = \frac{1}{|D|} \sum_{i=1}^{|D|} \Big( \mathbb{I}(S_{i,t}) - \mathbb{I}(S_{i,t-1}) \Big),
\end{equation}
where $ \mathbb{I}(\cdot) $ is an indicator function for correct answers. The final improvement after $ T $ turns becomes a proxy for the LLM's comprehensive ability:

\begin{equation}
\Delta P_T(S) = \sum_{t=1}^{T} \Delta P_t(S),
\end{equation}

and we average over all $ M $ student models to compute the Comprehensive Ability (CA):

\begin{equation}
CA = \frac{1}{M} \sum_{j=1}^{M} \Delta P_T(S_j).
\end{equation}

\subsection{Ability Taxonomy and Metrics}
\label{sec:ability_taxonomy}

% Building on Bloom’s taxonomy~\cite{krathwohl2002revision}, we categorize LLM abilities into four levels: Application, Judgment, Guidance, and Reflection. These are derived from the interaction dynamics within Teach2Eval.

By combining the guidance mechanism in this framework with the Bloom's taxonomy~\cite{krathwohl2002revision} of cognitive levels, we categorize the LLM's abilities into four levels: Application, Judgment, Guidance, and Reflection. In the Appendix~\ref{app:ability}, we demonstrate that, in this interactive setting, Comprehensive Ability can be naturally decomposed into Judgment Ability, Guidance Ability, and Reflection Ability.

% \textbf{Application Ability} is the accuracy of the LLM directly evaluate on the benchmark. This is the basic ability of the LLM to use its knowledge to provide answers. We denote \( AA \) as the Application Ability of the LLM when performing Zero-Shot testing. Formally, it can be expressed as:

\textbf{Application Ability (AA)} is the accuracy of the LLM directly evaluate on the benchmark. This is the basic ability of the LLM to use its knowledge to provide answers. We denote it as the accuracy of the LLM when performing Zero-Shot testing:

\begin{equation}
AA = \frac{1}{|D|} \sum_{i=1}^{|D|} \mathbb{I}(T(d_i)),
\end{equation}
where $ T(d_i) $ is the LLM’s direct answer to question $ d_i $.

% \textbf{Judgment Ability} refers to the LLM’s ability to effectively evaluate the correctness after receiving a question and its solution. Here we assume that when the student's final answer is correct, their solution is also correct. So we can use the gold answer to determine if the teacher model's judgment is correct. We denote \( JA \) as the accuracy of the LLM's judgment in the first turn, and define this ability as:

\textbf{Judgment Ability (JA)} refers to the LLM’s ability to effectively evaluate the correctness after receiving a question and its solution, we can use the gold answer to determine if the teacher model's judgment is correct. We denote it as the LLM’s correctness in evaluating students' initial responses:

\begin{equation}
JA = \frac{1}{M} \sum_{j=1}^{M} \frac{1}{|D|} \sum_{i=1}^{|D|} \mathbb{I}(J_1(d_i, a_{j,i,0})),
\end{equation}
where $ J_1(\cdot) $ is the LLM’s judgment in the first round.

\textbf{Guidance Ability (GA)} refers to the LLM’s capability to provide personalized guidance and corrections based on the solution of weak models after judging the given solution, quantifies how effectively the LLM improves student performance after incorrect initial answers:

% \textbf{Guidance Ability} refers to the LLM’s capability to provide personalized guidance and corrections based on the solution of weak models after judging the given solution. This ability enables the weak models to gradually refine their thought process and ultimately arrive at the correct answer. We denote \( GA \) as the accuracy of the LLM's judgment in the first turn, and define this ability as:

\begin{equation}
GA = \frac{1}{M} \sum_{j=1}^{M} \frac{\sum_{i=1}^{|D_{j,\text{incorrect}}|} \mathbb{I}(S_{j,i,1})}{|D_{j,\text{incorrect}}|}.
\end{equation}

\textbf{Reflection Ability (RA)} refers to the LLM's capability to engage in self-reflection across multiple turns of interaction. It captures the LLM’s capacity to refine its guidance across multiple rounds. For each turn $ t \geq 2 $:

\begin{equation}
RA_t = \frac{1}{|D_{c,t-1}|} \sum_{i=1}^{|D_{c,t-1}|} \Big[ \mathbb{I}(\neg S_{i,t-1} \land S_{i,t}) - \mathbb{I}(S_{i,t-1} \land \neg S_{i,t}) \Big],
\end{equation}
where $ D_{c,t-1} $ contains questions answered correctly in the previous round, and the overall Reflection Ability is the geometric mean of improvements:

\begin{equation}
RA = \frac{1}{M} \sum_{j=1}^{M} \left[ \prod_{t=2}^{T} (1 + RA_{j,t}) - 1 \right].
\end{equation}

\section{Experience}

\subsection{Experience Setup}
In order to expand the evaluation scope of our method as much as possible, we define the selection criteria for weak student models as follows: strong instruction-following abilities but weak original application abilities, the current mainstream end-side models all meet this condition. So we select four weak models: LLaMA3.2-1B~\cite{dubey2024llama}, Qwen2.5-1.5B~\cite{yang2024qwen2}, MiniCPM-2B~\cite{hu2024minicpm}, and InternLM2.5-1.8B~\cite{cai2024internlm2}. Additionally, we select 26 state-of-the-art models, including various families such as LLaMA, Qwen, and DeepSeek, for evaluation. Detailed information can be found in the Appendix~\ref{app:models}. The experiments are conducted using the VLLM 0.6.4 framework for inference, with the following settings: temperature set to 0.0, max\_token set to 2k, and the experiments were run on four H100 GPUs. We set the total number of turns to 3.

\begin{table}[htbp]
\centering
\caption{Comparison of Teach2Eval with Chatbot Arena and Livebench}
\label{tab:evaluation_comparison}
\begin{tabular}{lcc|cc}
\toprule
\multirow{2}{*}{\textbf{Evaluation Method}} & \multicolumn{2}{c}{\textbf{Chatbot Arena}} & \multicolumn{2}{c}{\textbf{Livebench}} \\ 
                                            & Kendall's Tau & Spearman Cor & Kendall's Tau & Spearman Cor \\ 
\midrule
Direct Evaluation                          & 0.581         & 0.775        & 0.538         & 0.709        \\ 
Teach2Eval                                 & \textbf{0.790}         & \textbf{0.911}        & \textbf{0.821}         & \textbf{0.923}         \\ 
\bottomrule
\end{tabular}
\end{table}

\subsection{Main Result}
We compare 26 models based on the experimental setup described above, and the detailed results in Appendix~\ref{app:result}. To verify the effectiveness of our evaluation, we compare our results with two prominent leaderboards. To ensure the consistency of the data distribution, we compare our reasoning scores with the Chatbot Arena Math benchmark and our overall scores with the Livebench Reasoning, Math, and Language benchmark. The correlation coefficients are shown in Table~\ref{tab:evaluation_comparison}. Compared with direct evaluation, our evaluation method achieves a higher correlation between two leaderboards, both above 0.90, and is also more cost-effective than two leaderboards.

% Table x in Appendix X shows the comparison between our results and these two leaderboards, and 

\begin{wrapfigure}{r}{0.5\textwidth} % 右对齐，宽度为页面宽度的一半
    \centering
    \includegraphics[width=0.5\textwidth]{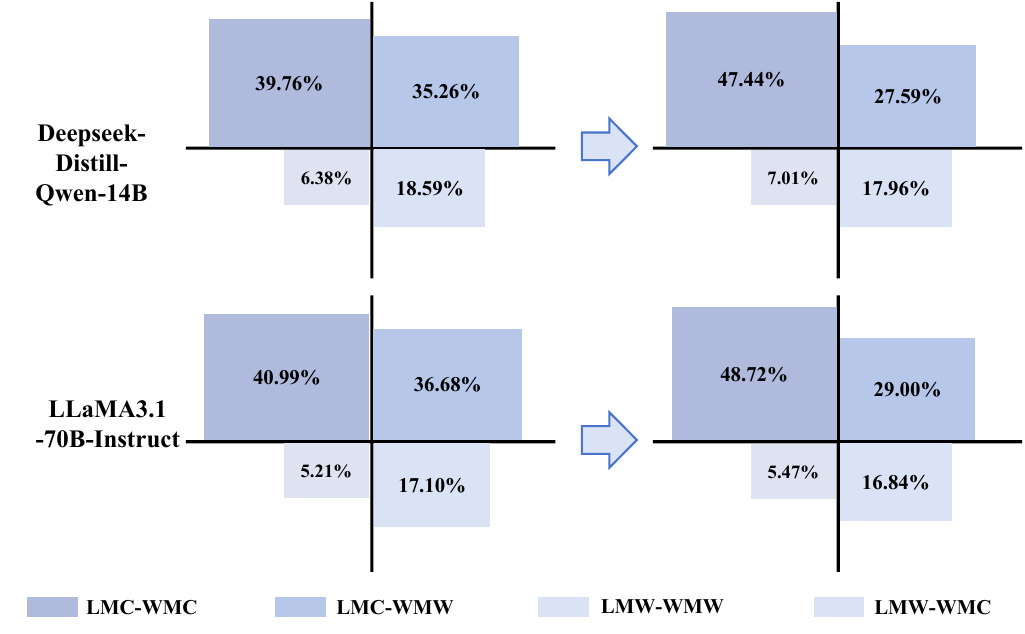}
    \caption{The confusion matrix comparison between DeepSeek and LLaMA models shows the left side without guidance and the right side after 3 turns of guidance. LM and WM represent LLM and weak models, while C and W represent correct and wrong models, respectively.}
    \label{img:Phi4}
\end{wrapfigure}

\begin{table*}[htbp]
    \centering
    \caption{The performance of the LLMs under Teach2eval demonstrates overall Comprehensive, Application, Evaluation, Guidance, and Reflection Ability, as well as Comprehensive Ability on four tasks.}
    \resizebox{\textwidth}{!}{%
    \begin{tabular}{lccccccccc}
        \toprule
        \multirow{3}{*}{Model} & \multicolumn{5}{c}{Overall} & \multirow{3}{*}{Knowledge} & \multirow{3}{*}{Reasoning} & \multirow{3}{*}{Understanding} & \multirow{3}{*}{Multilingual} \\
        \cmidrule(lr){2-6} 
                             & Comprehensive & Application & Judgment & Guidance & Reflection \\
                              & Ability        & Ability      & Ability  & Ability  & Ability     \\
        \midrule
        Qwen2.5-72B-Instruct                & 10.07 & 78.91 & 74.49 & 20.89 & 3.74 & 11.72 & 11.26 & 5.53 & 14.61 \\
        DeepSeek-V3                    & 9.84 & 75.20 & 73.78 & 19.45 & 5.05 & 10.29 & 11.12 & 5.71 & 15.07 \\
        DeepSeek-R1-Distill-Llama-70B       & 9.54 & 76.46 & 73.79 & 18.80 & 3.51 & 10.74 & 11.11 & 5.76 & 11.38 \\
        DeepSeek-R1-Distill-Qwen-32B        & 8.68 & 77.56 & 75.00 & 16.67 & 3.62 & 9.12 & 10.04 & 5.51 & 11.25 \\
        GPT-4o-0806                         & 8.55 & 76.97 & 75.70 & 15.60 & 6.86 & 9.73 & 9.96 & 4.71 & 10.77 \\
        DeepSeek-R1-Distill-Qwen-14B        & 8.31 & 75.03 & 72.52 & 16.71 & 2.91 & 7.86 & 10.32 & 5.12 & 10.20 \\
        Llama-3.3-70B-Instruct              & 7.97 & 77.69 & 75.05 & 13.27 & 5.74 & 8.61 & 9.44 & 5.19 & 8.53 \\
        Phi-4                               & 7.69 & 76.58 & 73.20 & 15.53 & 3.14 & 7.80 & 9.50 & 4.42 & 9.49 \\
        Qwen2.5-Coder-32B-Instruct          & 7.40 & 76.20 & 72.33 & 16.22 & 3.07 & 7.02 & 8.66 & 5.13 & 9.64 \\
        Llama-3.1-70B-Instruct              & 7.13 & 75.10 & 71.73 & 15.31 & 3.74 & 8.22 & 7.84 & 4.62 & 9.04 \\
        Qwen2.5-32B-Instruct                & 6.85 & 79.23 & 74.09 & 14.33 & 3.74 & 7.82 & 7.30 & 4.80 & 8.82 \\
        Qwen2-72B-Instruct                  & 6.16 & 74.20 & 72.45 & 13.83 & 3.43 & 7.11 & 6.35 & 4.89 & 6.98 \\
        DeepSeek-R1-Distill-Qwen-7B         & 5.61 & 61.93 & 53.82 & 12.53 & 1.67 & 3.48 & 8.68 & 2.62 & 6.31 \\
        Qwen2.5-14B-Instruct                & 5.49 & 76.38 & 71.13 & 12.88 & 2.73 & 6.60 & 5.76 & 4.03 & 6.11 \\
        Llama-3-70B-Instruct                & 4.17 & 73.03 & 70.89 & 10.36 & 2.30 & 5.09 & 3.78 & 4.02 & 4.14 \\
        Qwen2.5-7B-Instruct                 & 3.93 & 72.26 & 66.95 & 9.93 & 1.96 & 4.08 & 4.38 & 2.85 & 4.88 \\
        Yi1.5-34B-Chat                      & 3.93 & 68.61 & 61.74 & 8.93 & 2.17 & 3.31 & 4.44 & 3.46 & 4.52 \\
        DeepSeek-R1-Distill-Llama-8B        & 3.60 & 63.87 & 60.45 & 11.15 & 0.41 & 2.34 & 5.59 & 1.90 & 3.11 \\
        Yi1.5-9B-Chat                       & 2.56 & 64.19 & 63.55 & 12.05 & 0.87 & 2.66 & 2.82 & 2.06 & 2.68 \\
        InternLM2.5-20B                    & 2.16 & 64.94 & 62.20 & 8.12 & -0.03 & 2.14 & 2.25 & 2.19 & 1.67 \\
        Gemma-2-27b-it                 & 1.89 & 70.38 & 67.63 & 8.05 & 0.58 & 1.74 & 1.82 & 1.75 & 2.95 \\
        Llama-3.1-8B-Instruct               & 1.79 & 62.37 & 54.12 & 10.13 & 0.55 & 2.37 & 1.55 & 1.46 & 2.42 \\
        Gemma-2-9b-it                  & 1.70 & 65.63 & 64.77 & 7.79 & 0.40 & 1.43 & 1.57 & 1.46 & 3.52 \\
        Llama-3-8B-Instruct                 & 1.37 & 59.66 & 53.79 & 9.69 & 0.42 & 1.90 & 1.14 & 1.31 & 1.37 \\
        InternLM2.5-7B                     & 1.01 & 55.73 & 52.77 & 5.06 & -1.02 & 0.77 & 0.97 & 1.24 & 1.02 \\
        Yi1.5-6B-Chat                       & 0.91 & 57.12 & 54.08 & 9.34 & -0.85 & 1.67 & 0.82 & 0.28 & 1.40 \\
        \bottomrule
    \end{tabular}
    }
    \label{tab:model_comparison}
\end{table*}

As shown in Table~\ref{tab:model_comparison}, we find that Qwen2.5-72B-Instruct and DeepSeek-V3 performs the best across all four types of tasks, significantly outperforming the other models. The next best models are GPT-4o and DeepSeek-R1-Distill series, which show similar performance. DeepSeek-R1-Distill-Qwen-14B and Phi-4 has only 14B parameters, demonstrating that even models of this scale can achieve exceptional performance. For models with 8B parameters or smaller, DeepSeek-R1-Distill-Qwen-7B shows strong performance, especially in the reasoning tasks, where its ability even rivals that of the 70B variant from the same family. Due to DeepSeek-R1-Distill-Qwen-14B’s impressive performance with models below 20B, we compare its confusion matrix with that of Llama3.1-70B-Instruct, as shown in Figure \ref{img:Phi4}. The results indicate that DeepSeek-R1-Distill-Qwen-14B can teach weak models more effectively within the capacity of teacher models, showcasing its higher-level abilities.

\subsection{Ability Dimensional Analysis}
\begin{figure*}[htbp]
\centerline{\includegraphics[width=\textwidth]{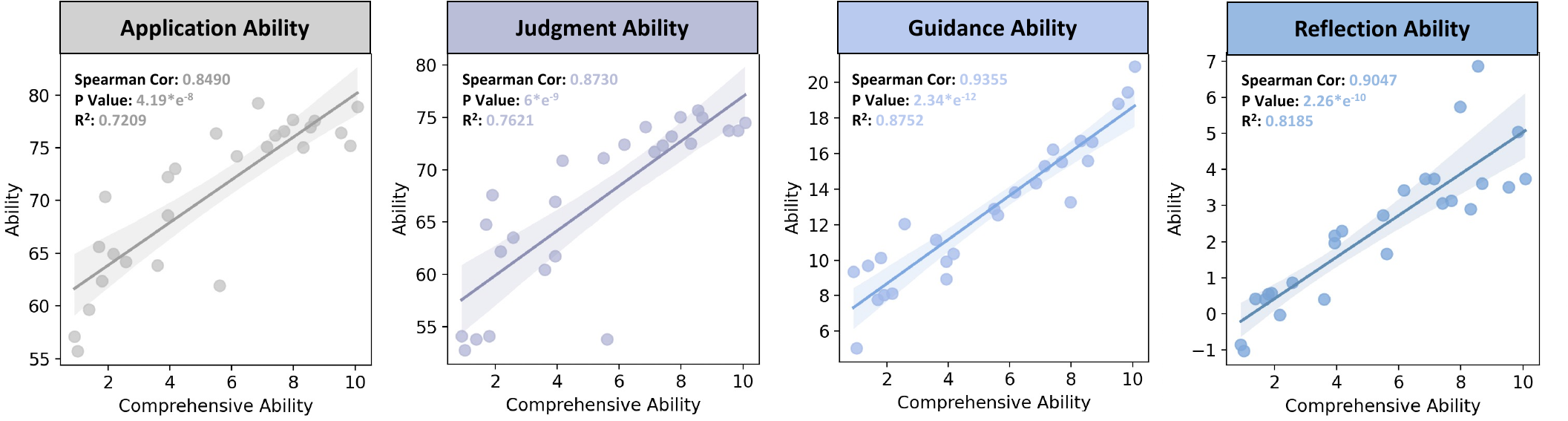}}
\caption{The correlation coefficient between Comprehensive Ability and the four abilities.}
\label{img:abilities}
\end{figure*}

To further explore the factors influencing the overall performance of LLMs, we use the classification method designed in Section \ref{sec:ability_taxonomy} to assess the four capabilities of all LLMs and compare them with their Comprehensive Ability. Table \ref{tab:model_comparison} presents a comparison of the four capabilities with the Comprehensive Ability across all datasets. As shown in Figure \ref{img:abilities}, we observe that the correlation between the Comprehensive Ability and the four capabilities gradually increases. The Application Ability obtained through traditional direct evaluation has a relatively low correlation with our Comprehensive Ability, while the correlation for the higher-order abilities is above 0.9. For some models that show anomalies in direct evaluations, such as Qwen2.5-32B-Instruct, which achieves the highest Application Ability, our evaluation method reveals its true capabilities.

Regarding Judgment Ability, the data for all models exceed 50\%, indicating that all LLMs possess judgment capabilities for weak models. However, there are significant differences among the models in the comparisons for Guidance and Reflection abilities. For Reflection Ability, the models with the worst performance, Yi-1.5-6B-Chat and InternLM2.5-7B, exhibits inconsistent reflection, performing worse than the other models.

\subsection{Ablation Experiment}

To verify the robustness and convergence of Teach2Eval, we conducted two ablation experiments.

% \textbf{Ablation1: There is no bias in the random selection of the weak model group.} To verify whether the selection of weak models would introduce bias to the evaluation results, we further investigate whether randomly removing a weak model from the group would lead to instability. In our setup, there are a total of four weak models, resulting in four combinations of three weak models. The correlations with Chatbot Arena are 0.907, 0.911, 0.911, and 0.907, respectively, and the correlations with LiveBench are all 0.92, confirming the robustness of the evaluation mechanism and the absence of bias.
\textbf{Ablation1: No bias in random selection of weak models.} To assess potential bias from weak model selection, we evaluate the stability of results when one of the four weak models is randomly removed. This yields four combinations of three models each. The resulting correlations with Chatbot Arena (0.907–0.911) and LiveBench (all 0.92) confirm the robustness and impartiality of the evaluation mechanism.

\begin{wrapfigure}{r}{0.5\textwidth} % 右对齐，宽度为页面宽度的一半
    \centering
    \includegraphics[width=1\linewidth]{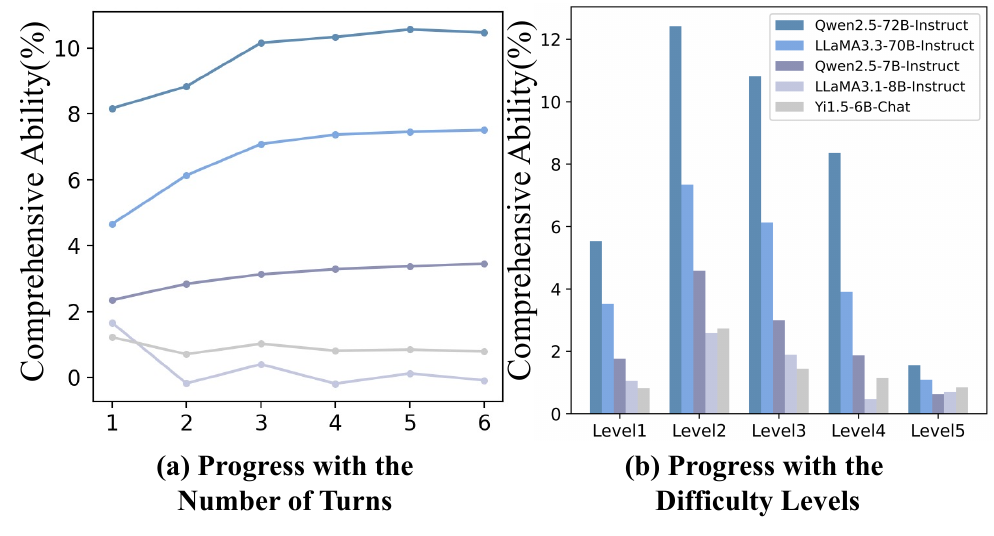}
    \caption{(a) The improvement of weak model capability with changes in the number of guiding turns; (b) The amount of improvement guided by different models on data of different difficulty levels.}
    \label{img:turns}
\end{wrapfigure}

\textbf{Ablation2: Weak model improvement tends to stabilize after multiple guidance turns.} To evaluate the convergence and Reflective Ability of LLMs, we select five representative models, each providing six rounds of guidance to weak models (performance shown in Figure~\ref{img:turns}(a)). While high-performing models show initial gains, the improvement generally plateaus as the number of rounds increases. Lower-performing models like LLaMA3.1-8B and Yi1.5-6B reach a plateau even earlier, indicating limited reflective capability. These results confirm the convergence behavior of our method. In our main experiments, we adopt the performance after three turns as the final metric, as most models have nearly converged by then.

% \textbf{Ablation2: After multiple turns of guidance, the improvement of the weak model tends to stabilize.} In order to demonstrate the convergence of the method and study the reflectivity of LLM, we select five representative models. Each model provides 6 turns of guidance to each weak model, and the performance is shown in Figure \ref{img:turns}(a).  For models with better performance, as the number of guide wheels increases, the improvement of weak models also gradually increases. However, for the lower-performing models such as LLaMA3.1-8B and Yi1.5-6B, the results reflect that this models have weak reflective ability and cannot continuously guide weak models  At the same time, This result also shows the discriminability and convergence of our method. Most models converge after 3 turns, and in the main experiment, we choose the effect of 3 turns as the final metric.

\section{Insights}
% Teach2Eval除了能给出准确的模型群体排名外，提供的细粒度指标正交于所有的benchmarks，能够将静态的task-specific评估转换为动态的capability-based评估，对于模型的分析也自然能够引导模型进一步训练与调整。

\begin{wrapfigure}{r}{0.5\textwidth} % 右对齐，宽度为页面宽度的一半
    \centering
    \includegraphics[width=1\linewidth]{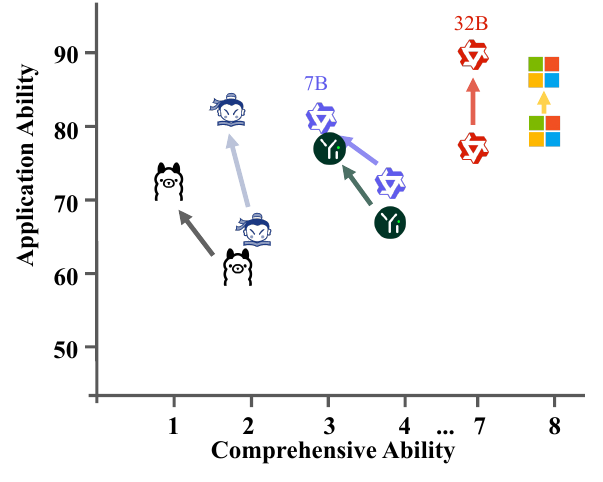}
    \caption{The trend of changes before and after fine-tuning six LLMs: models with strong abilities can maintain Comprehensive Ability, while models with weaker abilities show varying degrees of attenuation.}
    \label{img:contamination}
\end{wrapfigure}

In addition to providing effective model rankings, Teach2Eval can also offer fine-grained indicators orthogonal to exist benchmarks, which can transform static task-specific evaluations into dynamic capability-based evaluations, this analysis of the model can naturally guide further training and refinement.

\textbf{Insight1: Models affected by data contamination perform inadequately in higher-order capabilities.} Teach2Eval not only replicates data contamination but also prevents its occurrence during model training. To verify its effectiveness in addressing data contamination issue, we construct a subset of data for distillation and evaluation. Specifically, we used GPT-4o to traverse and answer the subset, then applied rejection sampling based on the correct answers and randomly filter 3,000 samples. Six models were fine-tuned on this subset, as illustrated in the Figure ~\ref{img:contamination}. The results show that most distilled models exhibit improved Application Ability but diminished Comprehensive Ability, whereas Qwen2.5-32B and Phi-4, stronger models, remain largely unchanged — demonstrating that Teach2Eval effectively mitigates evaluation issues caused by data contamination. Furthermore, tracking model performance across training stages within this coordinate system provides feedback that helps reduce the risk of overfitting or data contamination.

% Teach2eval不仅能够复制数据污染，其同样可以在模型训练的过程中防止模型出现数据污染问题。为了验证其能够解决数据污染问题，我们构建了部分数据用以蒸馏并进行测试。具体而言，利用GPT-4o对数据集进行遍历作答，根据正确答案使用拒绝采样方法随机过滤出3000条数据，并在这个集合上对六个模型进行微调，如图所示。可以发现绝大部分经过蒸馏的模型的Application Ability增强，但Comprehensive Ability减弱，只有能力较强的Qwen2.5-32B基本保持不变，这一现象能够验证Teach2Eval能够解决数据污染评测问题。同时，模型训练过程中可以观察其不同阶段在此坐标轴中的移动情况，这种反馈能够有效降低模型过拟合或数据污染的风险。

\begin{figure}[]
\centerline{\includegraphics[width=0.9\textwidth]{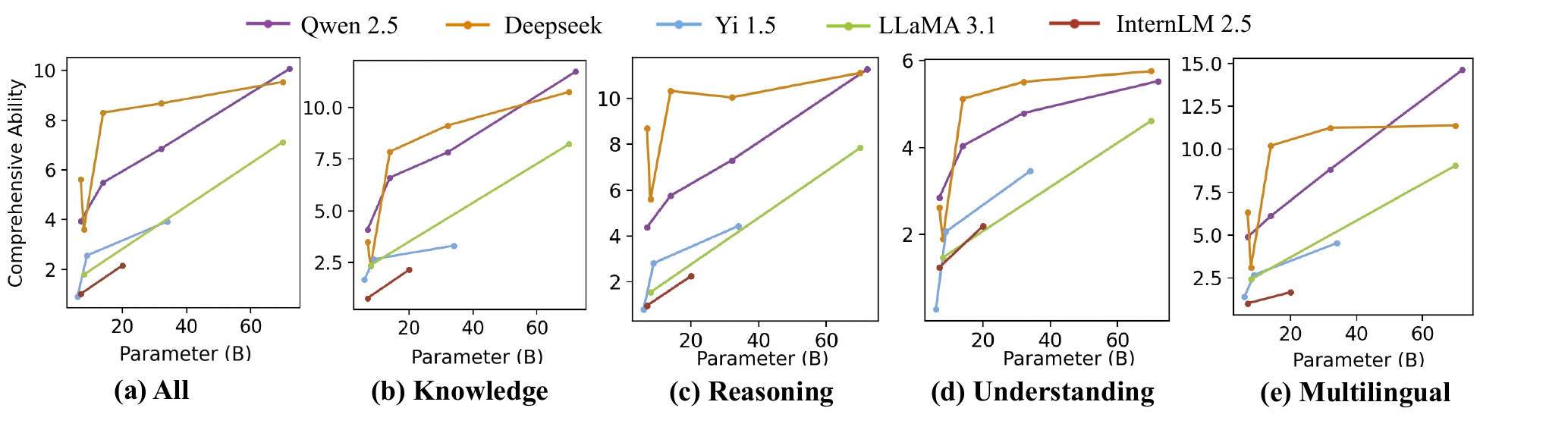}}
\caption{The trend of Comprehensive Ability changing with the parameters for model-families.}

\label{img:scaling_law}
\end{figure}

\textbf{Insight2: The Scaling Law remains valid for higher-order capabilities.} We conduct a study on the Scaling Law~\cite{kaplan2020scaling} of five model families, including Qwen, Deepseek, LLaMA, InternLM, and Yi, as shown in Figure \ref{img:scaling_law}. Our findings show that, across the overall dataset, the Comprehensive Ability increases with model size within each family. However, for the DeepSeek-Distill family models, variations in the size of the base models (Qwen or Llama) lead to fluctuations in their higher-order capabilities, which is more consistent with consensus. In terms of Application Ability, these models follow the trend that larger sizes yield better performance, which may result in incorrect evaluations during training, potentially leading to models perform poorly in reality. This suggests that while  models may appear to follow the Scaling Law in Application Ability, their true capabilities can differ, we can use higher-order capabilities to assess models' genuine performance.

\begin{wrapfigure}{r}{0.5\textwidth} % 右对齐，宽度为页面宽度的一半
    \centering
    \includegraphics[width=1\linewidth]{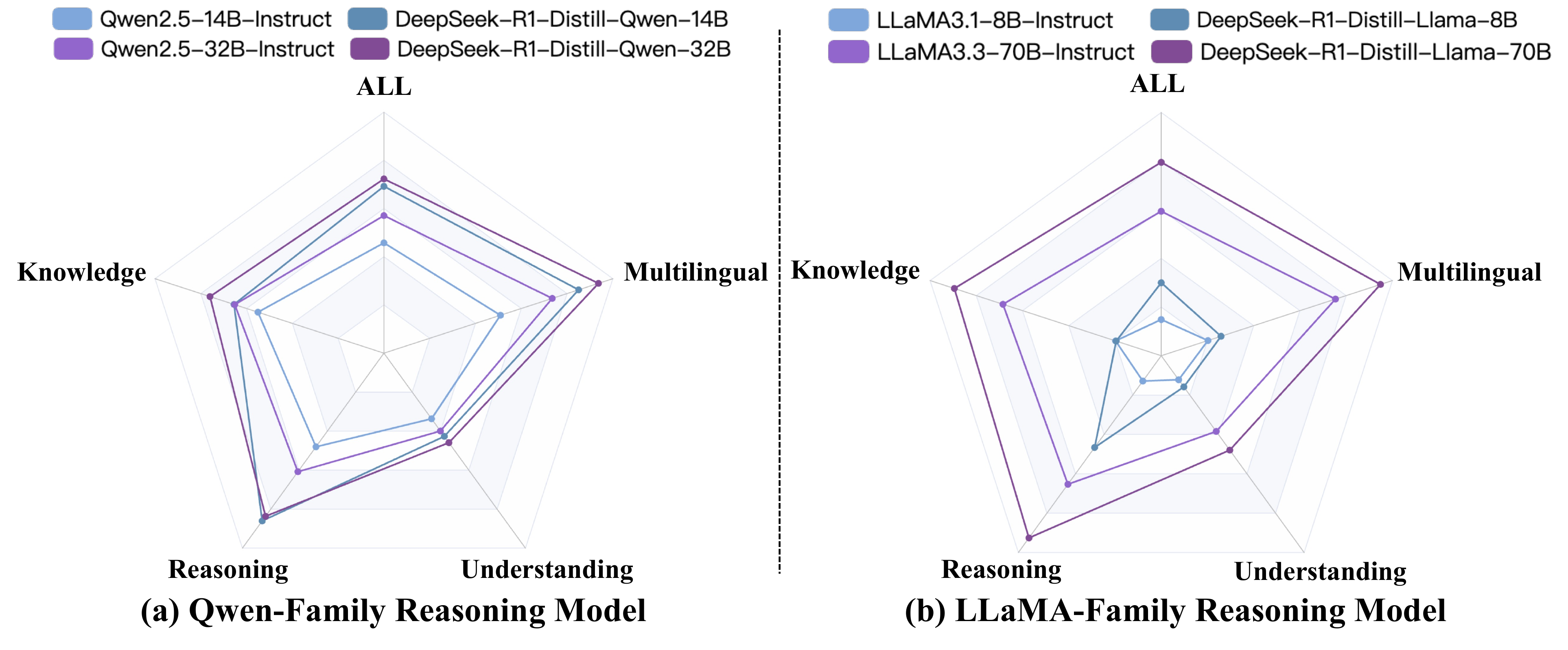}
    \caption{The Comprehensive Ability of reasoning LLMs and their base models varies on different types of tasks, and reasoning LLMs have shown improvements in various types, especially in reasoning tasks.}
    \label{img:reasoning}
\end{wrapfigure}

% 但对于DeepSeel-Distill 系列模型而言，不同大小的基模型不同导致了其高阶能力上存在波动，这是更符合认知的。但其在Application Ability中，完全符合参数越大能力越强的趋势，这可能会导致训练时的错误判定，得到了真实表现不佳的模型。 This suggests that while these models may follow the Scaling Law in terms of Application Ability, their true capabilities may differ，我们可以根据高阶能力来判断模型的真实表现。

\textbf{Insight3: Reasoning enhancement is effective in many tasks, but its impact varies across different tasks.} Currently, reasoning LLMs like OpenAI-o1~\cite{jaech2024openai} and Deepseek-R1~\cite{guo2025deepseek} are gaining popularity. These models use test-time scaling to further enhance their reasoning abilities, which has attracted significant attention. To assess the abilities of these reasoning models, we also select the Deepseek family, which were distilled based on Qwen and LLaMA models. In Figure~\ref{img:reasoning}, we compare the reasoning LLMs with their base models. Our findings show that the reasoning models outperform the base models across all tasks, where the 14B reasoning model surpasses the 32B base model, especially in the Reasoning and Multilingual tasks. The improvement in the Knowledge task is less pronounced, and only minimal gains are observed in the Understanding task.

\textbf{Insight4:The low-order capabilities of weak models can be leveraged to inversely evaluate the high-order capabilities of stronger models.} As models become increasingly capable, it becomes challenging to find stronger models or sufficiently broad datasets for effective evaluation. Previous research~\cite{khan2024debating} has explored whether the critical capabilities of weaker models can be used to judge the responses of stronger models; however, such subjective evaluations are inherently limited by the capacity ceiling of the weaker models. Teach2Eval leverages the response-generation capability of weak models, elevating the evaluation ceiling to match that of the stronger teacher models. This enables more accurate and effective assessment of model capabilities, making it particularly suitable for potential future AGI scenarios.

% 当模型能力越来越强后，我们可能难以找到更强的模型以及更广泛的数据集对其进行有效评估。过去的研究关注于探究【】能否利用弱模型的批判能力judge强模型的回答，但这种主观评价必然受弱模型的上限限制。Teach2Eval利用弱模型的回答能力，将评价的上限提升为大模型能力的上限，能够更准确有效的评估模型能力，适用于未来可能的AGI场景。

% \textbf{Insight5: Medium-difficulty problems are more successfully guided to improvements.} To analyze the relationship between guidance and model capability across data difficulty levels, we calculate the guidance improvement ratios for each difficulty set. The statistical information can be found in Appendix~\ref{app:dataset}, and we also present five models in Figure \ref{img:turns}(b). We find that the guidance effect is most effective on data of medium difficulty, and this effect exhibits a decreasing trend as the difficulty level increases. Strong models show robust guidance abilities across all difficulty levels. This indicates that further attention needs to be paid to the model's understanding of simple problems during model training.

\textbf{Insight5: Medium-difficulty problems are more successfully guided to improvements.} To analyze the relationship between guidance and model capability across data difficulty levels, we calculate the guidance improvement ratios for each difficulty set. The statistical information can be found in Appendix~\ref{app:dataset}, and we also present five models in Figure \ref{img:turns}(b). We find that the guidance effect is most effective on data of medium difficulty, and this effect exhibits a decreasing trend as the difficulty level increases. Strong models show robust guidance abilities across all difficulty levels. This indicates that further attention needs to be paid to the model's understanding of simple problems during model training, which may be overlooked in favor of more complex cases.

\section{Conclusion}
% To better assess the capabilities of LLMs, we draw inspiration from the Feynman Technology and design an evaluation method in which LLMs guide weak models, and the weak models provide answers. The improvement in the weak models' performance serves as an indicator of the LLMs' capabilities. We collect 60 datasets and design 15,000 questions, testing them across 25 models. The results show that our method exhibits a high degree of similarity to Chatbot Arena, LiveBench, and GPT-as-a-Judge, but at a fraction of the cost of direct evaluation. This approach offers an extensible, interpretable, and quantifiable evaluation mechanism.

% Teach2Eval offers a scalable approach to evaluating LLMs by leveraging their ability to guide weaker models, with performance improvements serving as a metric for the LLM's abilities. This method addresses the limitations of traditional evaluation metrics, providing a more cost-effective, accurate, and multidimensional assessment. Our experiments show that Teach2Eval aligns with existing platforms and remains robust across different scales, making it a promising tool for future LLM evaluations.

Teach2Eval presents a promising direction for evaluating LLMs in a way that is both scalable and cognitively meaningful. By reframing evaluation as a teaching process, we move beyond conventional performance metrics toward a more holistic understanding of model capabilities. The method naturally addresses key limitations in current practices—such as dataset bias, contamination, and lack of fine-grained insights—while remaining cost-effective and fully automated. Our experiments demonstrate that student-driven evaluation aligns well with established rankings and provides actionable feedback for model development. As LLMs continue to evolve, methods like Teach2Eval offer a practical and insightful path forward for fair, comprehensive, and interpretable evaluation.

\bibliography{anthology,custom}
\bibliographystyle{acl_natbib}

%%%%%%%%%%%%%%%%%%%%%%%%%%%%%%%%%%%%%%%%%%%%%%%%%%%%%%%%%%%%
\clearpage
\appendix

\section{Description of Ability Deconstruction}
\label{app:ability}

We use the improvement in weak model performance as an indicator of the LLM’s comprehensive ability. Assuming all abilities are for the same weak student model. This comprehensive ability is influenced by the three-tier capabilities of the LLM: judgment ability \( JA \), guidance ability \( GA \), and reflection ability \( RA \). By breaking down the comprehensive ability, we derive the relationships between these different capabilities.

First, let \( \Delta S_1 \) represent the improvement in the weak model's performance after the first turn of guidance, which can be expressed as follows:

\[
\Delta P_1(S) = JA_1' \cdot GA_1 \Bigl( JA_1 \cdot \neg P(S_0) + P(S_0) \cdot \neg JA_1 \Bigr) - \alpha P(S_0) \cdot \neg JA_1
\]

Where \(\neg P(S_0) \) and \(\neg JA_1 \) represent the probability of initial answer errors in the weak model and the probability of judgment errors in the LLM in first turn, respectively. \(\alpha \) is a fluctuation factor used to measure the conflict between the LLM's judgment and guidance, such as when the judgment is correct but the guidance is erroneous. \(JA'\) is influenced by the elements of \(JA\), and it is calculated as follows:

\[
J' = \frac{\sum_{d_i \in D}(\neg P(S_0(d_i)))}{\sum_{d_i \in D}(\neg JA_1(d_i))}
\]

Next, we consider the improvement in the weak model’s performance across multiple turns of guidance. Let \( \Delta S_T \) denote the performance improvement of the weak model at the \( T \)-th turn. Assuming that the LLM’s reflection ability \( RA \) continuously influences performance improvement in each turn, we can express the performance improvement after multiple turns of guidance as follows:

\[
\Delta P_T(S)  = \Delta P_1(S) \cdot (RA + 1)
\]

Through these equations, we describe that the comprehensive ability of the LLM is a result of the combined effects of its judgment, guidance, and reflection abilities. Specifically, the guidance and judgment abilities interact to improve the performance of the weak model, while the reflection ability continues to influence performance improvement across multiple turns of guidance.

\begin{figure}[htbp]
\centerline{\includegraphics[width=0.7\linewidth]{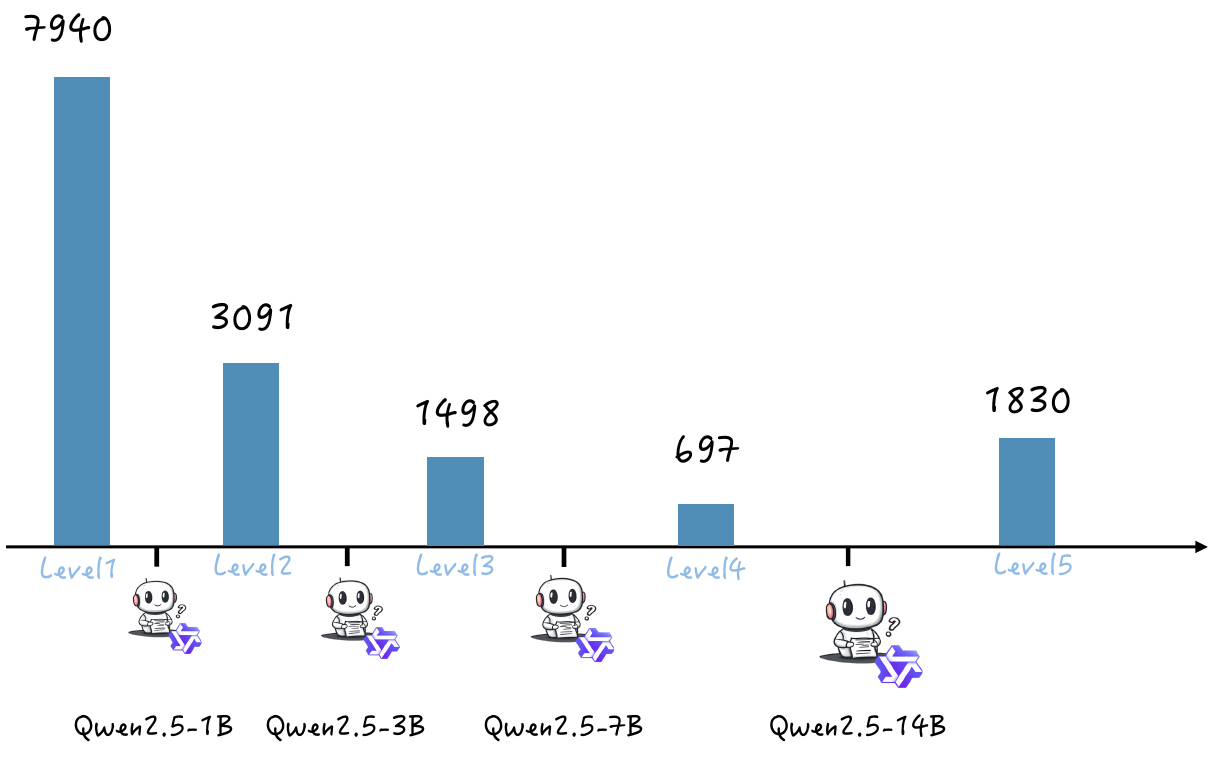}}
\caption{Use Qwen-family models for data difficulty classification.}
\label{img:data_diffi}
\end{figure}

\section{Datasets and Data Construction}
\label{app:dataset}
\begin{figure}[htbp]
\centerline{\includegraphics[width=1\linewidth]{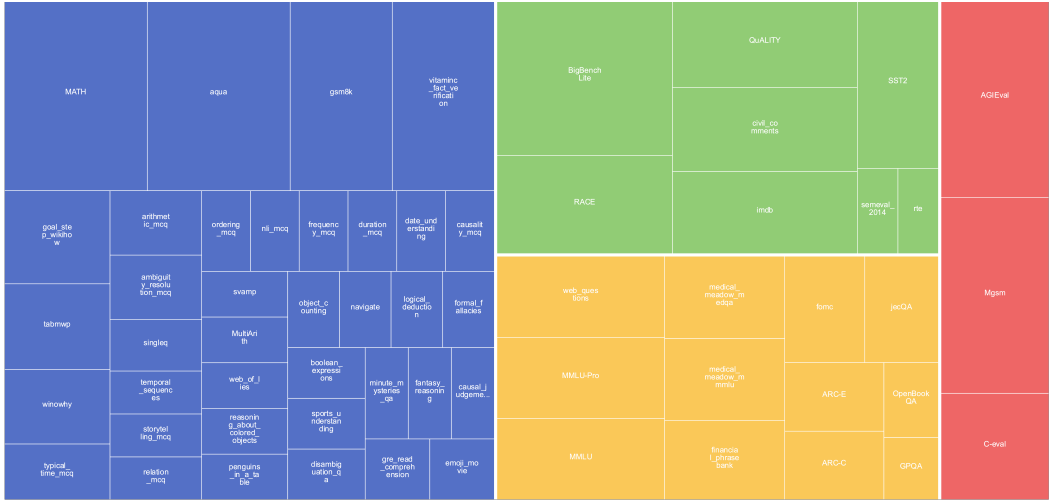}}
\caption{Dataset summary visualization, where blue represents Reasoning task, green represents Understanding task, yellow represents Knowledge task, red represents Multilingual task, and block size represents the number of samples in the dataset.}
\label{img:dataset}
\end{figure}

To ensure the comprehensiveness and effectiveness of our evaluation, we collect 60 datasets and sample 15,000 pieces of data, classify them into four tasks: Knowledge, Reasoning, Understanding, and Multilingual. Figure~\ref{img:dataset} is the data statistical chart, and Table~\ref{tab:dataset} is the data statistical table.

In order to modify all datasets to MCQ format, we use 10 weak models such as Qwen2.5-1.5B, Llama3.2-1B, etc., set the Temperature to 0.7, and randomly answer each question until we collected 3 incorrect answers. We use GPT-4o as the rewriter and reviewer, with gold answer as the correct answer and weak models answer as the incorrect answer for format conversion. We randomly place the correct answer positions during construction.

Afterwards, in order to classify all the data into difficulty categories, we used Qwen family models to answer each question sequentially from 1B to 14B. Classify the data into the difficulty category corresponding to the first correctly answered model, and set the question that all models cannot answer as the highest difficulty. Finally, divide it into five difficulty types, as shown in the Figure~\ref{img:data_diffi}.

\begin{table}[htbp]
\centering

\caption{Dataset statistical information}
\label{tab:dataset}
\begin{tabular}{c|cc|cc}
\hline
Task         & Name                   & Num. & Name                & Num. \\ \hline
\multirow{6}{*}{Knowledge}    & ARC-C                  & 202  & ARC-E               & 231  \\ 
                              & GPQA                   & 103 & MMLU                 & 433  \\
                              & MMLU-Pro               & 433  & OpenBookQA          & 115  \\
                              & web\_questions          & 433  & fomc                & 246  \\
                              & financial\_phrasebank   & 289  & jecQA               & 231  \\
                              & medical\_meadow\_mmmlu  & 289  & medical\_meadow\_medqa & 289  \\ \hline
\multirow{2}{*}{Multilingual} & AGIEval                & 578  & C-eval              & 311  \\
                              & Mgsm                   & 577  &                      &      \\ \hline
\multirow{4}{*}{Understanding} & BigBenchLite           & 1156 & civil\_comments      & 462  \\
                               & imdb                   & 462  & QuALITY             & 482  \\
                               & RACE                   & 1141 & rte                 & 64   \\
                               & semeval\_2014          & 64   & SST2                & 421  \\ \hline
\multirow{19}{*}{Reasoning}     & emoji\_movie           & 50   & gre\_read\_comprehension & 32  \\
                               & vitaminc\_fact\_verification & 879  & causal\_judgement   & 50   \\
                               & fantasy\_reasoning     & 50   & goal\_step\_wikihow  & 393  \\
                               & minute\_mysteries\_qa  & 50   & winowhy             & 393  \\
                               & disambiguation\_qa     & 57   & sports\_understanding & 57   \\
                               & boolean\_expressions    & 57   & formal\_fallacies    & 57   \\
                               & logical\_deduction      & 57   & navigate            & 57   \\
                               & object\_counting        & 57   & penguins\_in\_a\_table & 50  \\
                               & reasoning\_about\_colored\_objects & 57   & web\_of\_lies       & 57   \\
                               & aqua                   & 1156 & GSM8k               & 305  \\
                               & MATH                   & 694  & MultiArith          & 50   \\
                               & singleq                & 115  & svamp               & 69   \\
                               & tabmwp                 & 231  & ambiguity\_resolution\_mcq & 92  \\
                               & arithmetic\_mcq        & 92   & causality\_mcq      & 50   \\
                               & date\_understanding     & 57   & duration\_mcq       & 92   \\
                               & frequency\_mcq         & 92   & nli\_mcq            & 92   \\
                               & ordering\_mcq          & 92   & relation\_mcq       & 92   \\
                               & storytelling\_mcq      & 92   & temporal\_sequences & 57   \\
                               & typical\_time\_mcq     & 92   &                      &      \\ \hline
\end{tabular}
\end{table}

\section{Evaluation Models}
\label{app:models}

In order to test the performance of the current model on various tasks, we select 26 models for testing, including current state-of-the-art open source models such as LLaMA, Qwen, DeepSeek, Gemma, etc. The detailed information is shown in Table \ref{tab:models}.

\begin{table}[]
\centering
\caption{Models Information}
\begin{tabular}{ccccc}
\hline
\textbf{Model Name} & \textbf{Model size} & \textbf{Organization} & \textbf{Deployment method} \\
\hline
GPT-4o-0806 & Unknown & OpenAI & API \\
Deepseek-v3 & 671B & DeepSeek & API \\
Qwen2.5-72B-Instruct & 72B & Alibaba  & Local \\
Qwen2-72B-Instruct & 72B & Alibaba  & Local \\
Llama3.3-70B-Instruct & 70B & Meta & Local \\
Llama3.1-70B-Instruct & 70B & Meta & Local \\
Llama3-70B-Instruct & 70B & Meta & Local \\
DeepSeek-R1-Distill-Llama-70B & 70B & DeepSeek & Local \\
Yi-1.5-34B-Chat & 34B & 01AI & Local \\
Qwen2.5-32B-Instruct & 32B & Alibaba  & Local \\
Qwen2.5-Coder-32B-Instruct & 32B & Alibaba  & Local \\
DeepSeek-R1-Distill-Qwen-32B & 32B & DeepSeek & Local \\
Gemma-2-27b-it & 27B & Google & Local \\
InternLM2.5-20B & 20B & Shanghai AI Lab & Local \\
Qwen2.5-14B-Instruct & 14B & Alibaba  & Local \\
DeepSeek-R1-Distill-Qwen-14B & 14B & DeepSeek & Local \\
Phi-4 & 14B & Microsoft & Local \\
Yi-1.5-9B-Chat & 9B & 01AI & Local \\
Gemma2-9B-it & 9B & Google & Local \\
DeepSeek-R1-Distill-Llama-8B & 8B & DeepSeek & Local \\
Llama-3.1-8B-Instruct & 8B & Meta & Local \\
Llama-3-8B-Instruct & 8B & Meta & Local \\
Qwen2.5-7B-Instruct & 7B & Alibaba  & Local \\
DeepSeek-R1-Distill-Qwen-7B & 7B & DeepSeek & Local \\
InternLM2.5-7B & 7B & Shanghai AI Lab & Local \\
Yi-1.5-6B-Chat & 6B & 01AI & Local \\

\hline
\label{tab:models}
\end{tabular}
\end{table}

\section{Prompts}
\label{app:prompts}

\subsection{Weak Student Model Prompts}

\begin{figure}[htbp]
\centerline{\includegraphics[width=1\linewidth]{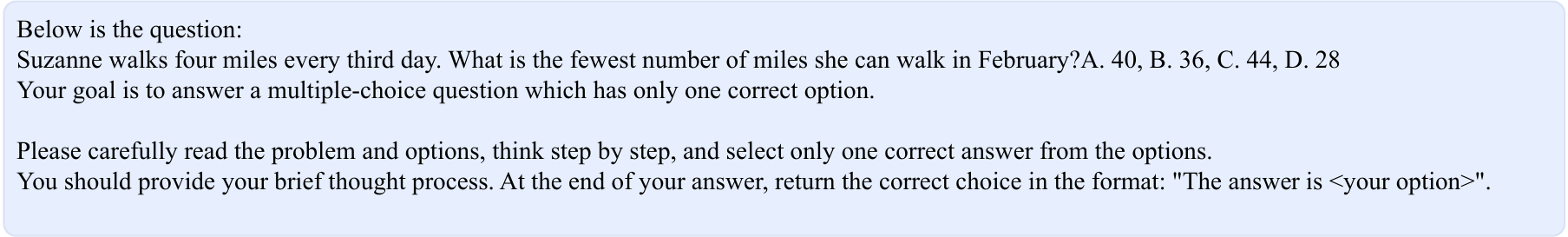}}
\caption{Prompt of weak model to answer questions directly.}
\label{img:weak_dir_prompt}
\end{figure}

\begin{figure}[htbp]
\centerline{\includegraphics[width=1\linewidth]{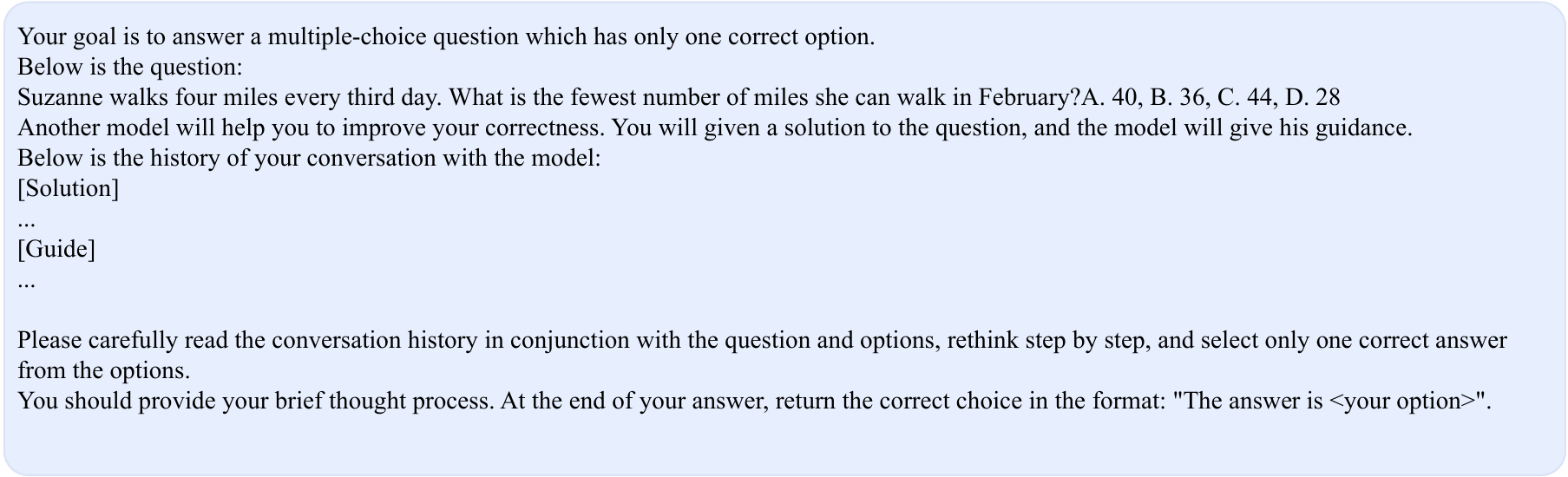}}
\caption{Prompt of weak model to re-answer questions according to the guide of the LLM.}
\label{img:weak_prompt}
\end{figure}

The main task of the weak model is to answer questions directly and to re-answer questions according to the guide of the LLM. We have designed two prompts for this purpose, as shown in Figure ~\ref{img:weak_dir_prompt} and Figure ~\ref{img:weak_prompt}.

\subsection{Teacher LLM Prompts}

\begin{figure}[htbp]
\centerline{\includegraphics[width=1\linewidth]{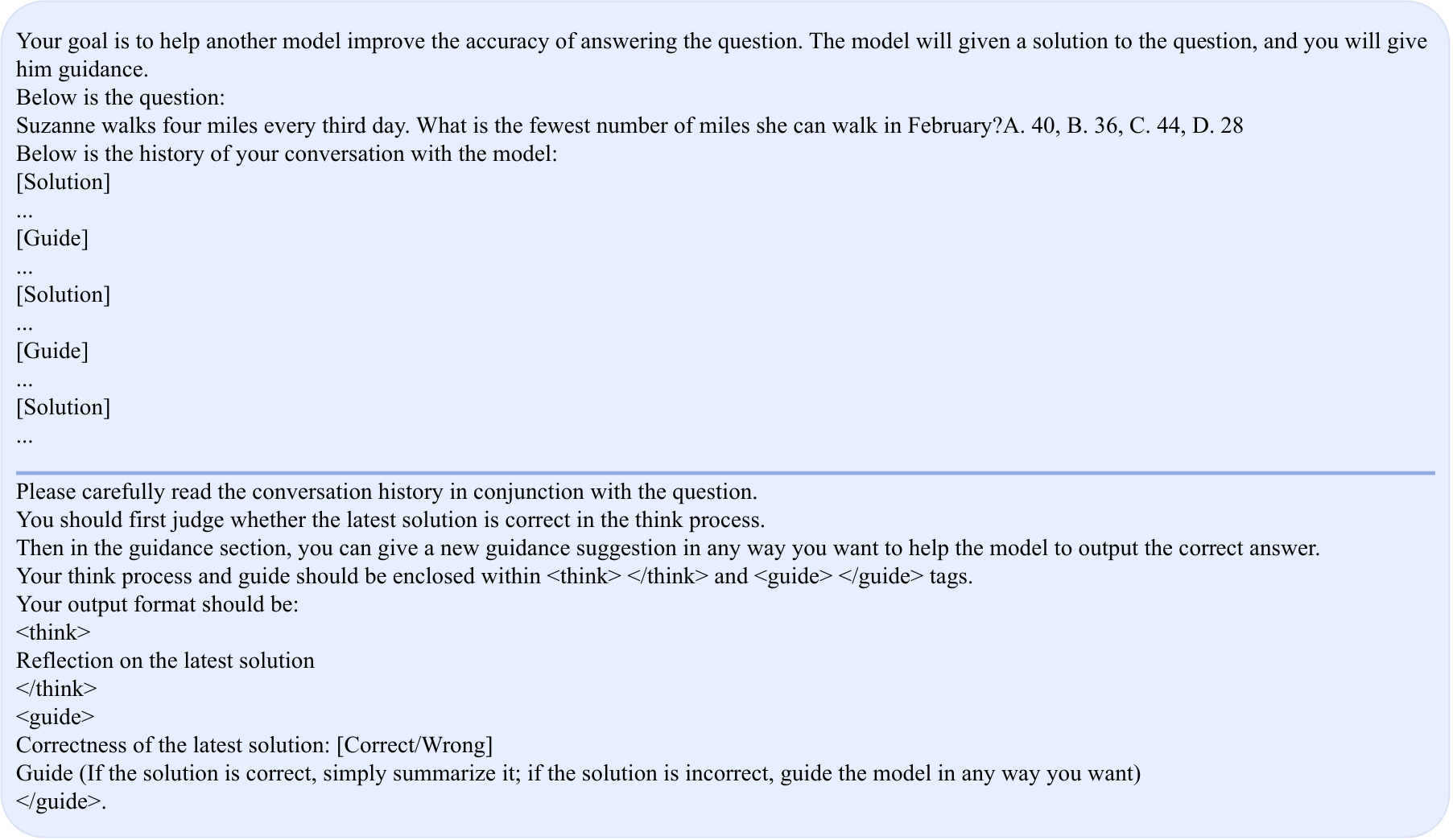}}
\caption{Prompt of LLM to judge and guide based on history solutions and guidances.}
\label{img:teacher_prompt}
\end{figure}

We have designed prompts for teacher models, aimed at enabling them to reflect better and generate better guides based on the past teacher-student conversations, as shown in Figure ~\ref{img:teacher_prompt}

\subsection{LLM-as-a-Judge Prompt}

\begin{figure}[htbp]
\centerline{\includegraphics[width=1\linewidth]{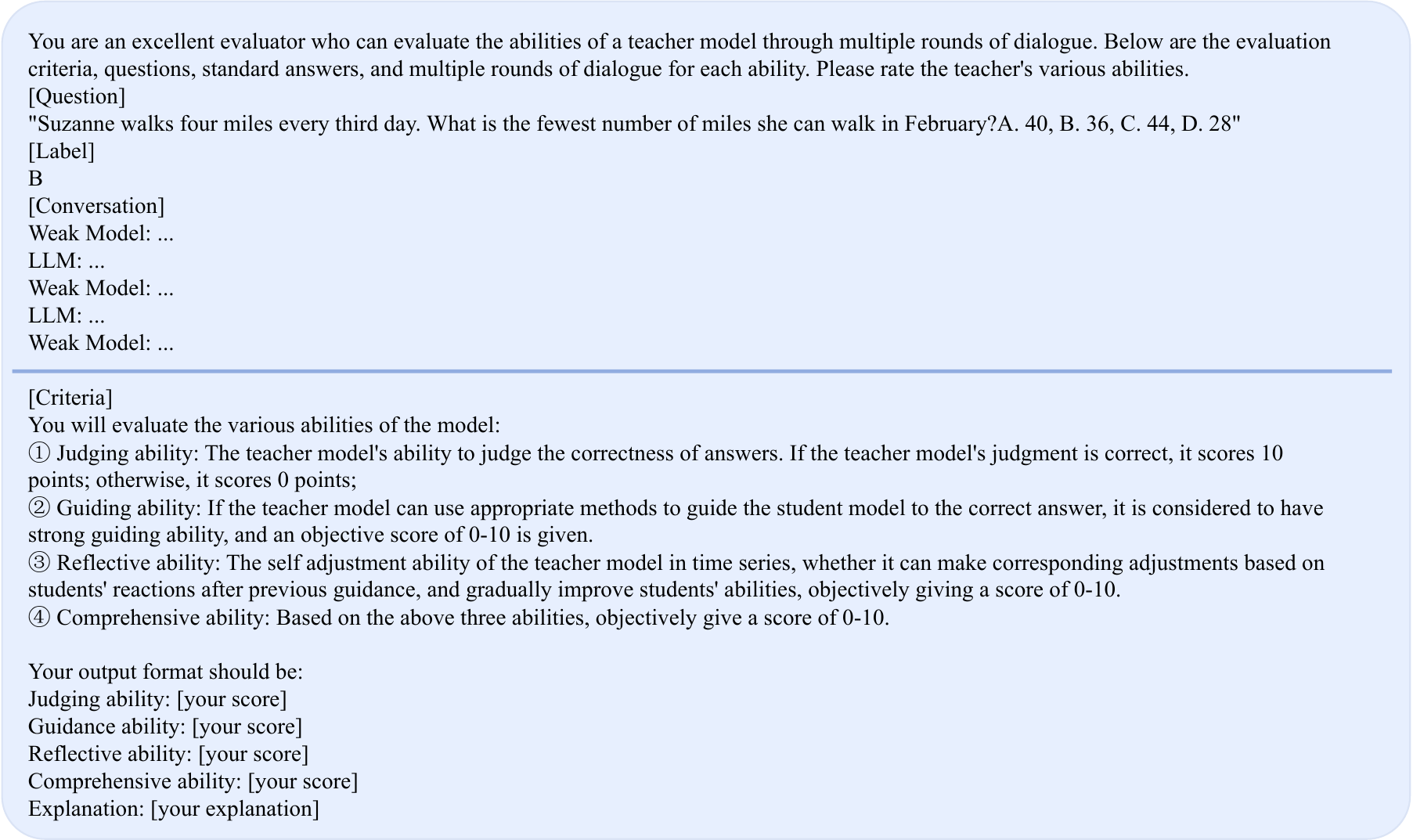}}
\caption{Prompt of LLM-as-a-Judge.}
\label{img:judge_prompt}
\end{figure}

We also design criteria for LLM-as-a-Judge, Figure ~\ref{img:judge_prompt} shows the prompt using a LLM as a judge.

\section{Test Result}
\label{app:result}

In addition to the overall abilities presented in the main text, we also analyze the performance of all models on various abilities for each task, including Knowledge, Reasoning, Understanding, and Multilingual. Refer to Table~\ref{tab:knowledge}, Table~\ref{tab:reasoning}, Table~\ref{tab:understanding}, and Table~\ref{tab:multilingual} for details.

\begin{table}[htbp]
    \centering
    \caption{Knowledge Task Performance}
    \label{tab:knowledge}
    \small % Reduce font size
    \begin{tabularx}{\textwidth}{l *{5}{>{\centering\arraybackslash}X}} % Use tabularx for flexible width
        \toprule
        Model Name & CA & AA & JA & GA & RA \\
        \midrule
        Qwen2.5-72B-Instruct & 11.72 & 74.90 & 77.66 & 19.05 & 5.30 \\
        DeepSeek-R1-Distill-Llama-70B & 10.74 & 69.03 & 73.78 & 16.21 & 5.36 \\
        DeepSeek-V3 & 10.29 & 68.11 & 76.98 & 17.06 & 5.89 \\
        GPT-4o-0806 & 9.73 & 74.25 & 78.29 & 13.49 & 10.14 \\
        DeepSeek-R1-Distill-Qwen-32B & 9.12 & 71.42 & 74.82 & 13.73 & 5.15 \\
        Llama-3.3-70B-Instruct & 8.61 & 71.90 & 77.22 & 9.50 & 10.06 \\
        Llama-3.1-70B-Instruct & 8.22 & 67.81 & 73.91 & 12.56 & 5.92 \\
        DeepSeek-R1-Distill-Qwen-14B & 7.86 & 68.33 & 70.89 & 12.90 & 2.72 \\
        Qwen2.5-32B-Instruct & 7.81 & 72.97 & 76.28 & 11.97 & 5.61 \\
        Phi-4 & 7.80 & 72.36 & 75.60 & 11.84 & 4.92 \\
        Qwen2-72B-Instruct & 7.11 & 67.87 & 75.09 & 11.01 & 5.49 \\
        Qwen2.5-Coder-32B-Instruct & 7.02 & 66.92 & 74.22 & 12.19 & 4.34 \\
        Qwen2.5-14B-Instruct & 6.60 & 70.96 & 73.88 & 10.89 & 4.74 \\
        Llama-3-70B-Instruct & 5.09 & 68.39 & 72.12 & 8.50 & 4.15 \\
        Qwen2.5-7B-Instruct & 4.08 & 62.18 & 65.56 & 7.49 & 3.19 \\
        DeepSeek-R1-Distill-Qwen-7B & 3.48 & 47.32 & 60.10 & 7.36 & 0.31 \\
        Yi1.5-34B-Chat & 3.31 & 66.52 & 58.80 & 6.10 & 1.84 \\
        Yi1.5-9B-Chat & 2.66 & 57.99 & 66.45 & 8.60 & 0.19 \\
        Llama-3.1-8B-Instruct & 2.37 & 53.90 & 59.82 & 7.04 & 1.00 \\
        DeepSeek-R1-Distill-Llama-8B & 2.34 & 53.78 & 60.06 & 7.08 & -0.12 \\
        InternLM2.5-20B & 2.14 & 62.06 & 58.88 & 5.95 & 0.32 \\
        Meta-Llama-3-8B-Instruct & 1.90 & 55.18 & 59.94 & 6.53 & 1.56 \\
        Gemma-2-27b-it & 1.74 & 64.05 & 71.48 & 4.94 & 1.55 \\
        Yi1.5-6B-Chat & 1.66 & 50.50 & 59.65 & 6.46 & -0.16 \\
        Gemma-2-9b-it & 1.43 & 59.58 & 68.18 & 4.74 & 0.75 \\
        InternLM2.5-7B & 0.77 & 55.43 & 49.38 & 3.27 & -1.02 \\
        \bottomrule
    \end{tabularx}
\end{table}

\begin{table}[htbp]
    \centering
    \caption{Reasoning Task Performance}
    \label{tab:reasoning}
    \small % Reduce font size
    \begin{tabularx}{\textwidth}{l *{5}{>{\centering\arraybackslash}X}} % Use tabularx for flexible width
        \toprule
        Model Name & CA & AA & JA & GA & RA \\
        \midrule
        Qwen2.5-72B-Instruct             & 11.26 & 83.44 & 74.08 & 24.73 & 4.12 \\
        DeepSeek-V3                    & 11.12 & 78.80 & 73.23 & 24.54 & 5.10 \\
        DeepSeek-R1-Distill-Llama-70B      & 11.11 & 81.18 & 75.61 & 23.93 & 3.04 \\
        DeepSeek-R1-Distill-Qwen-14B     & 10.32 & 79.96 & 76.60 & 22.44 & 3.20 \\
        DeepSeek-R1-Distill-Qwen-32B     & 10.04 & 82.35 & 77.44 & 21.41 & 3.37 \\
        GPT-4o-0806                      & 9.96 & 79.13 & 75.39 & 19.96 & 6.32 \\   
        Phi-4                           & 9.50 & 82.81 & 74.82 & 20.63 & 2.80 \\
        Llama-3.3-70B-Instruct           & 9.44 & 84.31 & 75.48 & 17.42 & 5.76 \\   
        DeepSeek-R1-Distill-Qwen-7B     & 8.68 & 71.67 & 51.21 & 18.57 & 2.81 \\
        Qwen2.5-Coder-32B-Instruct          & 8.66 & 82.32 & 71.86 & 20.49 & 3.05 \\
        Llama-3.1-70B-Instruct            & 7.84 & 81.36 & 72.30 & 19.00 & 4.07 \\  
        Qwen2.5-32B-Instruct            & 7.30 & 84.43 & 73.90 & 17.36 & 3.44 \\
        Qwen2-72B-Instruct               & 6.35 & 79.40 & 71.58 & 16.72 & 2.84 \\     
        Qwen2.5-14B-Instruct            & 5.76 & 82.65 & 70.74 & 15.19 & 2.65 \\
        DeepSeek-R1-Distill-Llama-8B    & 5.59 & 71.82 & 62.81 & 15.83 & 1.43 \\
        Yi1.5-34B-Chat                  & 4.44 & 72.68 & 63.97 & 11.36 & 2.24 \\
        Qwen2.5-7B-Instruct             & 4.38 & 79.09 & 67.25 & 12.54 & 1.97 \\    
        Llama-3-70B-Instruct            & 3.78 & 77.11 & 70.76 & 11.95 & 1.65 \\    
        Yi1.5-9B-Chat                 & 2.82 & 70.24 & 62.69 & 15.66 & 0.94 \\
        InternLM2.5-20B                & 2.25 & 63.68 & 63.10 & 9.93 & 0.12 \\     
        Gemma-2-27b-it                   & 1.82 & 74.41 & 65.91 & 9.57 & 0.78 \\
        Gemma-2-9b-it                   & 1.57 & 68.38 & 63.57 & 9.27 & 0.49 \\    
        Llama-3.1-8B-Instruct        & 1.54 & 68.51 & 51.73 & 12.76 & 0.55 \\            
        Llama-3-8B-Instruct         & 1.14 & 60.96 & 51.29 & 12.46 & 0.02 \\        
        InternLM2.5-7B             & 0.97 & 50.61 & 54.59 & 6.05 & -1.13 \\
        Yi1.5-6B-Chat                & 0.82 & 61.95 & 51.70 & 12.04 & -0.31 \\  
        \bottomrule
    \end{tabularx}
\end{table}

\begin{table}[htbp]
    \centering
    \caption{Understanding Task Performance}
    \label{tab:understanding}
    \small % Reduce font size
    \begin{tabularx}{\textwidth}{l *{5}{>{\centering\arraybackslash}X}} % Use tabularx for flexible width
        \toprule
        Model Name & CA & AA & JA & GA & RA \\
        \midrule
        DeepSeek-R1-Distill-Llama-70B    & 5.76 & 75.35 & 70.72 & 14.43 & 2.03 \\        
        DeepSeek-V3                    & 5.71 & 74.48 & 70.29 & 14.29 & 2.44 \\
        Qwen2.5-72B-Instruct            & 5.53 & 75.16 & 72.10 & 16.09 & 1.13 \\
        DeepSeek-R1-Distill-Qwen-32B    & 5.51 & 75.09 & 71.10 & 12.33 & 2.22 \\
        Llama-3.3-70B-Instruct          & 5.19 & 73.33 & 72.40 & 11.50 & 2.58 \\
        Qwen2.5-Coder-32B-Instruct         & 5.13 & 75.31 & 70.57 & 13.89 & 1.52 \\
        DeepSeek-R1-Distill-Qwen-14B     & 5.12 & 73.19 & 67.72 & 12.11 & 1.80 \\  
        Qwen2-72B-Instruct              & 4.89 & 70.98 & 70.99 & 13.06 & 2.18 \\
        Qwen2.5-32B-Instruct            & 4.79 & 76.34 & 72.18 & 12.72 & 1.97 \\
        GPT-4o-0806                       & 4.71 & 77.52 & 73.15 & 12.19 & 3.72 \\  
        Phi-4                           & 4.42 & 73.05 & 68.97 & 11.68 & 1.90 \\
        Llama-3.1-70B-Instruct          & 4.62 & 72.53 & 68.94 & 13.34 & 0.93 \\
        Qwen2.5-14B-Instruct            & 4.04 & 72.18 & 68.58 & 12.30 & 1.06 \\
        Llama-3-70B-Instruct            & 4.02 & 72.22 & 70.06 & 10.94 & 1.64 \\
        Yi1.5-34B-Chat                  & 3.46 & 66.44 & 60.70 & 8.50 & 1.80 \\
        Qwen2.5-7B-Instruct             & 2.85 & 71.19 & 67.36 & 9.18 & 0.44 \\
        DeepSeek-R1-Distill-Qwen-7B    & 2.62 & 61.01 & 49.62 & 9.03 & 0.31 \\
        InternLM2.5-20B                 & 2.19 & 71.38 & 63.71 & 8.63 & -0.61 \\
        Yi1.5-9B-Chat                   & 2.06 & 62.30 & 61.59 & 11.65 & 0.64 \\
        DeepSeek-R1-Distill-Llama-8B    & 1.90 & 64.79 & 56.61 & 8.98 & -0.51 \\
        Gemma-2-27b-it                  & 1.75 & 70.93 & 66.06 & 9.14 & -0.40 \\
        Llama-3.1-8B-Instruct           & 1.46 & 63.17 & 50.11 & 10.25 & 0.01 \\   
        Gemma-2-9b-it                     & 1.46 & 67.83 & 62.82 & 8.44 & -0.31 \\                      
        Llama-3-8B-Instruct          & 1.31 & 64.09 & 49.47 & 9.98 & 0.07 \\       
        InternLM2.5-7B             & 1.24 & 67.05 & 53.77 & 5.93 & -1.21 \\
        Yi1.5-6B-Chat             & 0.28 & 57.46 & 50.40 & 8.79 & -2.06 \\ 
        \bottomrule
    \end{tabularx}
\end{table}

\begin{table}[htbp]
    \centering
    \caption{Multilingual Task Performance}
    \label{tab:multilingual}
    \small % Reduce font size
    \begin{tabularx}{\textwidth}{l *{5}{>{\centering\arraybackslash}X}} % Use tabularx for flexible width
        \toprule
        Model Name & CA & AA & JA & GA & RA \\
        \midrule
        DeepSeek-V3                             & 15.07 & 78.45 & 78.54 & 19.01 & 11.62 \\
        Qwen2.5-72B-Instruct                    & 14.61 & 80.00 & 76.02 & 22.31 & 6.90 \\
        DeepSeek-R1-Distill-Llama-70B            & 11.38 & 76.72 & 75.15 & 16.56 & 6.84 \\
        DeepSeek-R1-Distill-Qwen-32B            & 11.25 & 78.63 & 76.57 & 16.37 & 6.34 \\
        GPT-4o-0806                             & 10.77 & 72.73 & 78.20 & 13.04 & 12.89 \\
        DeepSeek-R1-Distill-Qwen-14B            & 10.20 & 74.88 & 73.17 & 15.64 & 5.79 \\
        Qwen2.5-Coder-32B-Instruct      & 9.64 & 74.20 & 75.10 & 15.31 & 5.82 \\
        Phi-4                           & 9.49 & 70.44 & 73.38 & 14.52 & 5.13 \\
        Llama-3.1-70B-Instruct          & 9.04 & 72.90 & 72.56 & 12.99 & 7.41 \\
        Qwen2.5-32B-Instruct            & 8.82 & 80.07 & 75.49 & 12.47 & 7.48 \\
        Llama-3.3-70B-Instruct              & 8.53 & 75.84 & 76.11 & 11.04 & 7.36 \\
        Qwen2-72B-Instruct                  & 6.98 & 76.18 & 74.42 & 11.47 & 6.51 \\
        DeepSeek-R1-Distill-Qwen-7B         & 6.31 & 56.86 & 62.80 & 10.90 & 3.55 \\
        Qwen2.5-14B-Instruct                & 6.11 & 74.74 & 74.04 & 10.42 & 4.81 \\
        Qwen2.5-7B-Instruct                 & 4.88 & 69.56 & 67.61 & 7.89 & 5.30 \\
        Yi1.5-34B-Chat                  & 4.52 & 62.73 & 62.06 & 7.59 & 4.12 \\
        Llama-3-70B-Instruct            & 4.15 & 68.81 & 71.04 & 7.54 & 4.09 \\
        Gemma-2-9b-it                   & 3.52 & 61.37 & 67.80 & 7.65 & 1.84 \\
        DeepSeek-R1-Distill-Llama-8B        & 3.11 & 50.85 & 62.65 & 8.75 & -0.43 \\
        Gemma-2-27b-it                  & 2.95 & 66.21 & 70.65 & 6.96 & 1.07 \\
        Yi1.5-9B-Chat                   & 2.68 & 58.43 & 66.38 & 7.84 & 3.39 \\
        Llama-3.1-8B-Instruct           & 2.42 & 53.58 & 62.90 & 7.43 & 1.55 \\
        InternLM2.5-20B                 & 1.67 & 57.95 & 61.48 & 5.34 & 0.63 \\
        Yi1.5-6B-Chat                   & 1.40 & 50.92 & 62.17 & 7.52 & -0.74 \\
        Llama-3-8B-Instruct             & 1.37 & 51.47 & 62.99 & 6.26 & 1.24 \\
        InternLM2.5-7B                  & 1.02 & 44.78 & 49.93 & 3.49 & 0.46 \\
        \bottomrule
    \end{tabularx}
\end{table}

\section{Discuss}
\subsection{Compare with GPT-as-a-Judge}

To further validate the reliability of Teach2Eval, we also compare it with the current mainstream LLM-as-a-Judge evaluation approach. We select 10 models and randomly pick 5000 dialogue samples for each model. Using GPT-4o, we score the dialogues on four abilities, with the evaluation criteria detailed in Appendix ~\ref{app:prompts}. We compare the Teach2Eval results with the scores from GPT and find a Spearman correlation of 0.883, while the correlation between Direct Evaluation and GPT scores is 0.833. Our method demonstrates a higher correlation.

\begin{figure}[]
\centerline{\includegraphics[width=0.9\linewidth]{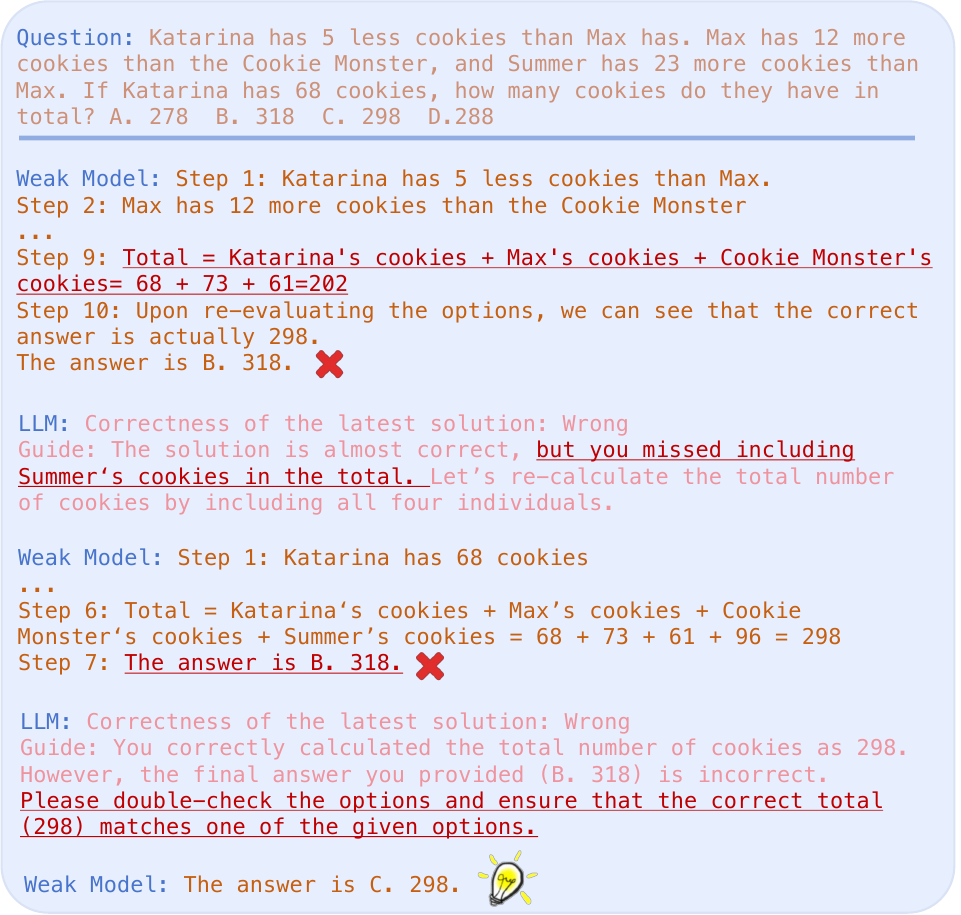}}
\caption{The case of the guidance process, where the weak model gradually answers in the correct direction under the guidance of the LLM.}
\label{img:demo}
\end{figure}

\subsection{Case Study}
To explore the effectiveness of LLMs in guiding weak models, we check several samples. Figure \ref{img:demo} presents one example. Upon observation, we find that high-performing models are able to identify the issues in the weak models' current responses and gradually guide them in the right direction, thereby improving the weak models' capabilities.

\section{Limitations}
Due to resource constraints and the length limitations of this article, we currently focus on evaluating large language models in the helpful domain. However, our method can easily be extended to other domains as well. Additionally, this approach can be applied to the multimodal domain, which we consider as part of our future work.

%%%%%%%%%%%%%%%%%%%%%%%%%%%%%%%%%%%%%%%%%%%%%%%%%%%%%%%%%%%%

\clearpage
\newpage

\end{document}